\documentclass{article}
\usepackage[nonatbib, preprint]{neurips_2021}
\usepackage[utf8]{inputenc}
\usepackage[T1]{fontenc}    
\usepackage[inline]{enumitem}
\usepackage{microtype}
\usepackage{graphicx}
\usepackage{booktabs}
\usepackage{multirow}
\usepackage{hyperref}
\usepackage{url}

\usepackage{caption}
\usepackage{mathptmx}
\usepackage{amssymb}
\usepackage{amsmath}

\usepackage[numbers]{natbib}
\usepackage{float}
\usepackage{caption}
\usepackage{subcaption}
\usepackage{amsfonts}       
\usepackage{nicefrac}       
\usepackage{microtype}      
\usepackage{xcolor} 
\usepackage{wrapfig}
\usepackage{verbatim}
\newcommand*\samethanks[1][\value{footnote}]{\footnotemark[#1]}

\title{Characterizing and Measuring the Similarity of Neural Networks with Persistent Homology}

\author{
    David Pérez-Fernández\thanks{Contributed equally.}\\
    SEGITTUR\\
    \texttt{david.perez@inv.uam.es}\\
    \And
    Asier Gutiérrez-Fandiño\samethanks\\
    Barcelona Supercomputing Center\\
    \texttt{asier.gutierrez@bsc.es}\\
    \And
    Jordi Armengol-Estapé\\
    Barcelona Supercomputing Center\\
    \texttt{jordi.armengol@bsc.es}\\
    \And
    Marta Villegas\\
    Barcelona Supercomputing Center\\
    \texttt{marta.villegas@bsc.es}\\
}

\begin{document}
\maketitle

\begin{abstract}
Characterizing the structural properties of neural networks is crucial yet poorly understood, and there are no well-established similarity measures between networks. In this work, we observe that neural networks can be represented as abstract simplicial complex and analyzed using their topological 'fingerprints' via Persistent Homology (PH). We then describe a PH-based representation proposed for characterizing and measuring similarity of neural networks. We empirically show the effectiveness of this representation as a descriptor of different architectures in several datasets. This approach based on Topological Data Analysis is a step towards better understanding neural networks and serves as a useful similarity measure.
\end{abstract}

\section{Introduction}
\setcounter{footnote}{0}
Machine learning practitioners can train different neural networks for the same task. Even for the same neural architecture, there are many hyperparameters, such as the number of neurons per layer or the number of layers. Moreover, the final weights for the same architecture and hyperparameters can vary depending on the initialization and the optimization process itself, which is stochastic. Thus, there is no direct way of comparing neural networks accounting for the fact that neural networks solving the same task should be measured as being similar, regardless of the specific weights. This also prevents one from finding and comparing modules inside neural networks (e.g., determining if a given sub-network does the same function as other sub-network in another model). Moreover, there are no well-known methods for effectively characterizing neural networks.

This work aims to characterize neural networks such that they can be measured to be similar once trained for the same task, with independence of the particular architecture, initialization, or optimization process. We focus on Multi-Layer Perceptrons (MLPs) for the sake of simplicity. We start by observing that we can represent a neural network as a directed weighted graph to which we can associate certain topological concepts.\footnote{See \citet{Jonsson2007SimplicialCO} for a complete reference on graph topology.} Considering it as a simplicial complex, we obtain its associated Persistent Diagram. Then, we can compute distances between Persistent Diagrams of different neural networks.

The proposed experiments aim to show that the selected structural feature, Persistent Homology, serves to relate neural networks trained for similar problems and that such a comparison can be performed by means of a predefined measure between the associated Persistent Homology diagrams. To test the hypothesis, we study different classical problems (MNIST, Fashion MNIST, CIFAR-10, and language identification and text classification datasets), different architectures (number and size of layers) as well as a control experiment (input order).

In summary, the main contributions of this work are the following:
\begin{itemize}
    \item We propose an effective graph characterization strategy of neural networks based on Persistent Homology.
    \item Based on this characterization, we suggest a similarity measure of neural networks.
    \item We provide empirical evidence that this Persistent Homology framework captures valuable information from neural networks and that the proposed similarity measure is meaningful.
\end{itemize}

The remainder of this paper is organized as follows. In Section \ref{sec:related}, we go through the related work. Then, in Section \ref{sec:method} we describe our proposal and the experimental framework to validate it. Finally, in sections \ref{sec:results} and  \ref{sec:conclusions} we report and discuss the results and arrive to conclusions, respectively.

\section{Related Work} 
\label{sec:related}
One of the fundamental papers of Topological Data Analysis (TDA) is presented in \citet{Carlsson2009TopologyAD} and suggests the use of Algebraic Topology to obtain qualitative information and deal with metrics for large amounts of data. For an extensive overview of simplicial topology on graphs, see \citet{Giblin1977GraphsSA, Jonsson2007SimplicialCO}. \citet{Aktas2019PersistenceHO} provide a thorough analysis of PH methods. 

More recently, a number of publications have dealt with the study of the capacity of neural networks using PH. \citet{Guss2018OnCT} characterize learnability of different neural architectures by computable measures of data complexity. \citet{Rieck2019NeuralPA} introduce the \textit{neural persistence} metric, a complexity measure based on TDA on weighted stratified graphs. This work suggests a representation of the neural network as a multipartite graph and the filtering of the Persistent Homology diagrams are performed for each layer independently. As the filtration contains at most 1-simplices (edges), they only capture zero-dimensional topological information, i.e. connectivity information. \citet{Donier2019CapacityAA} propose the concept of \textit{spatial capacity allocation analysis}. \citet{Konuk2019AnES} propose an empirical study of how NNs handle changes in topological complexity of the input data. 

In terms of pure neural network analysis, there are relevant works, like \citet{Hofer2020TopologicallyDD}, that study topological regularization. \citet{Clough2020ATL} introduce a method for training neural networks for image segmentation with prior topology knowledge, specifically via Betti numbers. \citet{Corneanu2020ComputingTT} try to estimate (with limited success) the performance gap between training and testing via neuron activations and linear regression of the Betti numbers.

On the other hand, topological analysis of decision boundaries has been a very prolific area. \citet{Ramamurthy2019TopologicalDA} propose a labeled Vietoris-Rips complex to perform PH inference of decision boundaries for quantification of the complexity of neural networks.

\citet{Naitzat2020TopologyOD} experiment on the PH of a wide range of point cloud input datasets for a binary classification problems to see that NNs transform a topologically rich dataset (in terms of Betti numbers) into a topologically simpler one as it passes through the layers. They also verify that the reduction in Betti numbers is significantly faster for ReLU activations than hyperbolic tangent activations.

\citet{Liu2020GeometryAT} obtain certain geometrical and topological properties of decision regions for neural models, and provide some principled guidance to designing and regularizing them. Additionally, they use curvatures of decision boundaries in terms of network weights, and the rotation index theorem together with the Gauss-Bonnet-Chern theorem.

Regarding neural network representations, one of the most related works to ours, \citet{Gebhart2019CharacterizingTS}, focuses on topological representations of neural networks. They introduce a method for computing PH over the graphical activation structure of neural networks, which provides access to the task-relevant substructures activated throughout the network for a given input. 

Interestingly, in \citet{Watanabe2020TopologicalMO}, authors work on neural network representations through simplicial complexes based on deep Taylor decomposition and they calculate the PH of neural networks in this representation. In \citet{Chowdhury2019PathHO}, they use directed homology to represent MLPs. They show that the path homology of these networks is non-trivial in higher dimensions and depends on the number and size of the network layers. They investigate homological differences between distinct neural network architectures. 

As far as neural network similarity measures are concerned, the literature is not especially prolific. In \citet{DBLP:journals/corr/abs-1905-00414}, authors examine similarity measures for representations (meaning, outputs of different layers) of neural networks based on canonical correlation analysis. However, note that this method \textit{compares neural network representations (intermediate outputs), not the neural networks themselves}. Remarkably, in \citet{7280769}, authors \textit{do} deal with the intrinsic similarity of neural networks themselves based on Forward Bipartite Alignment. Specifically, they propose an algorithm for aligning the topological structures of two neural networks. Their algorithm finds optimal bipartite matches between the nodes of the two MLPs by solving the well-known graph cutting problem. The alignment enables applications such as visualizations or improving ensembles. However, the methods only works under very restrictive assumptions,\footnote{For example, the two neural networks "must have the same number of units in each of their corresponding layers", and the match is done layer by layer.} and this line of work does not appear to have been followed up.

Finally, we note that there has been a considerable growth of interest in applied topology in the recent years. This popularity increase and the development of new software libraries,\footnote{\url{https://www.math.colostate.edu/~adams/advising/appliedTopologySoftware/}} along with the growth of computational capabilities, have empowered new works. Some of the most remarkable libraries are Ripser \cite{ctralie2018ripser, bauer2021ripser}, and Flagser \cite{Ltgehetmann2019ComputingPH}. They are focused on the efficient computation of PH. For GPU-Accelerated computation of Vietoris-Rips PH, Ripser++ \cite{Zhang2020GPUAcceleratedCO} offers an important speedup. The Python library we are using, Giotto-TDA \cite{tauzin2020giottotda}, makes use of both above libraries underneath.

We have seen that there is a trend towards the use of algebraic topology methods for having a better understanding of phenomena of neural networks and having more principled deep learning algorithms. Nevertheless, little to no works have proposed neural network characterizations or similarity measures based on intrinsic properties of the networks, which is what we intend to do.

\section{Methodology}
\label{sec:method}
In this section, we propose our method, which is heavily based on concepts from algebraic topology. We refer the reader to the Supplementary Material for the mathematical definitions. In this section, we also describe the conducted experiments.

Intrinsically characterizing and comparing neural networks is a difficult, unsolved problem. First, the network should be represented in an object that captures as much information as possible and then it should be compared with a measure depending on the latent structure. Due to the stochasticity of both the initialization and training procedure, networks are parameterized differently. For the same task, different functions that effectively solve it can be obtained. Being able to compare the trained networks can be helpful to detect similar neural structures.

We want to obtain topological characterizations associated to neural networks trained on a given task. For doing so, we use the Persistence Homology (from now on, PH) of the graph associated to a neural network. We compute the PH for various neural networks learned on different tasks. We then compare all the diagrams for each one of the task.

More specifically, for each of the studied tasks (image classification on MNIST, Fashion MNIST and CIFAR-10; language identification, and text classification on the Reuters dataset),\footnote{For more details, see Section \ref{sec:datasets}.} we proceed as follows:
\begin{itemize}
    \item We train several neural network models on the particular problem.
    \item We create a directed graph from the weights of the trained neural networks (after changing the direction of the negative edges and normalising the weights of the edges).
    \item We consider the directed graph as a simplicial complex and calculate its PH, using the weight of the edges as the filtering parameter, which range from 0 to 1. This way we obtain the so-called Persistence Diagram.
    \item We compute the distances between the Persistence Diagrams (prior discretization of the Persistence Diagram so that it can be computed) of the different networks.
    \item Finally, we analyze the similarity between different neural networks trained for the same task, for a similar task, and for a completely different task, independently of the concrete architecture, to see whether there is topological similarity.
\end{itemize}

As baselines, we set two standard matrix comparison methods that are the 1-Norm and the Frobenius norm. Having adjacency matrix $A$ and $B$, we compute the difference as $norm(A-B)$. However, these methods only work for matrices of similar size and thus, they are not general enough. We could also have used the Fast Approximate Quadratic assignment algorithm suggested in \citet{10.1371/journal.pone.0121002}, but for large networks this method becomes unfeasible to compute.


\subsection{Proposal}

Our method is as follows. We start by associating to a neural network a weighted directed graph that is analyzed as an abstract simplicial complex consisting on the union of points, edges, triangles, tetrahedrons and larger dimension polytopes (those are the elements referred as simplices). Abstract simplicial complexes are used in opposition to geometric simplicial complexes, generated by a point cloud embedded in the Euclidean space $\mathbb{R}^n$.

Given a trained neural network, we take the collection of neural network parameters as directed and weighted edges that join neurons, represented by graph nodes. Biases are considered as new vertices that join target neurons with an edge having a given weight. Note that, in this representation, we lose the information about the activation functions, for simplicity and to avoid representing the network as a multiplex network. Bias information could also have been ignored because we want large PH groups that characterize the network, while these connections will not change the homology group dimension of any order.

For negative edge weights, we reverse edge directions and maintain the absolute value of the weights. We discard the use of weight absolute value since neural networks are not invariant under weight sign transformations. This representation is consistent with the fact that every neuron can be replaced by a neuron from which two edges with opposite weights emerge and converge again on another neuron with opposite weights. From the point of view of homology, this would be represented as a closed cycle.

We then normalize the weights of all the edges as expressed in Equation \ref{formula:normalization} where $w$ is the weight to normalize, $W$ are all the weights and $\zeta$ is an smoothing parameter that we set to 0.000001. This smoothing parameter is necessary as we want to avoid normalized weights of edges to be 0. This is because 0 implies a lack of connection.

\begin{equation}
\label{formula:normalization}
max(1 - \frac{|w|}{max(|max(W)|, |min(W)|)}, \zeta)
\end{equation}

Given this weighted directed graph, we then define a directed flag complex associated to it. Topology of this directed flag complex can be studied using homology groups $H_n$. In this work we calculate homology groups up to degree 3 ($H_0$-$H_3$) due to computational complexity and our neural network representation method's layer connectivity limit. 

The dimensions of these homology groups are known as Betti numbers. The $i$-th Betti number is the number of $i$-dimensional voids in the simplicial complex ($\beta_0$ gives the number of connected components of the simplicial complex, $\beta_1$ gives the number of non reducible loops and so on).
For a deeper introduction to algebraic topology and computational topology, we refer to \citet{Edelsbrunner2009ComputationalT, Ghrist2014ElementaryAT}.

We work with a family of simplicial complexes, $K_\varepsilon$, for a range of values of $\varepsilon \in \mathbb{R}$ so that the complex at step $\varepsilon_t$ is embedded in the complex at $\varepsilon_{t+1}$ for $\varepsilon_t \leq \varepsilon_{t+1}$, i.e. $K_{\varepsilon} \subseteq K_{\varepsilon_{t+1}}$. In our case, $\varepsilon$ is the minimum weight of included edges of our graph representation of neural networks.

The nested family of simplicial complexes is called a \textit{filtration}. We calculate a sequence of homology groups by varying the $\varepsilon$ parameter, obtaining a persistence homology diagram. PH calculations are performed on $\mathbb{Z}_2$.

This filtration gives a collection of contained directed weighted graph or simplicial complex $K_{\varepsilon_{min}} \subseteq \ldots \subseteq K_{\varepsilon_t} \subseteq K_{\varepsilon_{t+1}} \subseteq \ldots \subseteq K_{\varepsilon_{max}}$, where $t \in [0,1]$ and $\varepsilon_{min} = 0$, $\varepsilon_{max} = 1$ (recall that edge weights are normalized).

Given a filtration, one can look at the birth, when a homology class appears, and death, the time when the homology class disappears. The PH treats the birth and the death of these homological features in $K_\varepsilon$ for different $\varepsilon$ values. Lifespan of each homological feature can be represented as an interval $(birth, death)$, of the homological feature. Given a filtration, one can record all these intervals by a Persistence Barcode (PB) \cite{Carlsson2009TopologyAD}, or in a Persistence Diagram (PD), as a collection of multiset of intervals.

As mentioned previously, our interest in this work is to compare PDs from two different simplicial complexes. There are two distances traditionally used to compare PDs, Wasserstein distance and Bottleneck distance. Their stability with respect to perturbations on PDs has been object of different studies \cite{Chazal2012PersistenceSF, CohenSteiner2005StabilityOP}.

In order to make computations feasible and to obviate noisy intervals, we filter the PDs by limiting the minimum PD interval size. We do so by setting a minimum threshold $\eta = 0.01$. Intervals with a lifespan under this value are not considered. Additionally, for computing distances, we need to remove infinity values. As we are only interested in the deaths until the maximum weight value, we replace all the infinity values by $1.0$.

Wasserstein distance calculations are computationally hard for large PDs (each PD of our NN models has a million persistence intervals per diagram). Therefore we use a vectorized version of PDs instead, also called PD discretization. This vectorized version summaries have been proposed and used on recent literature \cite{Adams2017PersistenceIA, Berry2020FunctionalSO, Bubenik2015StatisticalTD, Lawson2019PersistentHF, Rieck2019TopologicalML}. 

For the persistence diagram distance calculation, we use the Giotto-TDA library \cite{tauzin2020giottotda} and compute the following supported vectorized persistence summaries:
\begin{enumerate*}
    \item Persistence landscape.
    \item Weighted silhouette.
    \item Heat vectorizations.
\end{enumerate*}

\subsection{Experimental Framework}
\paragraph{Datasets}
\label{sec:datasets}
To determine the topological structural properties of trained NNs, we select different kinds of datasets. We opt for four well-known benchmarks in the machine learning community and one regarding language identification:
\begin{enumerate*}[label={(\arabic*)}]
\item the MNIST\footnote{\url{http://yann.lecun.com/exdb/mnist/}} dataset for classifying handwritten digit images,
\item the Fashion MNIST \cite{xiao2017/online} dataset for classifying clothing images into 10 categories,
\item the CIFAR-10\footnote{\url{https://www.cs.toronto.edu/~kriz/cifar.html}} (CIFAR) dataset for classifying 10 different objects,
\item the Reuters dataset for classifying news into 46 topics,
and \item the Language Identification Wikipedia dataset\footnote{\url{https://www.floydhub.com/floydhub/datasets/language-identification/1/data}} for identifying 7 different languages.
\end{enumerate*}

We selected these datasets because, apart from being well-known benchmarks, the performances without transfer learning are good enough and they have different data types and sizes. For CIFAR-10 and Fashion MNIST datasets we train a Convolutional Neural Network (CNN) first, and the convolutional layers are shared between all the models of the same dataset as a feature extractor. Recall that in this work we are focusing on MLPs, so we do not consider that convolutional weights. For the MNIST, Reuters and Language Identification datasets, we use an MLP. For Reuters and Language identification datasets, we vectorize the sentences with character frequency.

\paragraph{Experiments Pipeline}

We study the following variables (hyperparameters):
\begin{enumerate*}
    \item Layer width,
    \item Number of layers,
    \item Input order\footnote{Order of the input features, the control experiment. This one should definitely \textit{not} affect the performance in the neural networks, so if our method is correct, it should be uniform as per the proposed topological distances.}),
    \item Number of labels (number of considered classes).
\end{enumerate*}

We define the \textit{base} architecture as the one with a layer width of 512, 2 layers, the original features order, and considering all the classes (10 in the case of MNIST, Fashion MNIST and CIFAR, 46 in the case of Reuters and 7 in the case of the language identification task). Then, doing one change at a time, keeping the rest of the base architecture hyperparameters, we experiment with architectures with the following configurations:
\begin{itemize}
    \item \textbf{Layer width}: 128, 256, 512 (\textit{base}) and 1024.
    \item \textbf{Number of layers}: 2 (\textit{base}), 4, 6, 8 and 10.
    \item \textbf{Input order}: 5 different randomizations (with \textit{base} structure), the control experiment.
    \item \textbf{Number of labels} (MNIST, Fashion MNIST, CIFAR-10): 2, 4, 6, 8 and 10 (\textit{base}).
    \item \textbf{Number of labels} (Reuters): 2, 6, 12, 23 and 46 (\textit{base}).
    \item \textbf{Number of labels} (Language Identification): 2, 3, 4, 6 and 7 (\textit{base}).
\end{itemize}
Note that this is \textit{not} a grid search over all the combinations. We always modify one hyperparameter at a time, and keep the rest of them as in the base architecture. In other words, we experiment with all the combinations such that only one of the hyperparameters is set to a non-base value at a time.

For each dataset, we train 5 times (each with a different random weight initialization) each of these neural network configurations. Then, we compute the topological distances (persistence landscape, weighted silhouette, heat) among the different architectures. In total, we obtain $5 \times 5 \times 3$ distance matrices (5 datasets, 5 random initializations, 3 distance measures). Finally, we average the 5 random initializations, such that we get $5 \times 3$ matrices, one for each distance on each dataset. All the matrices have dimensions $19 \times 19$, since 19 is the number of experiments for each dataset (corresponding to the total the number of architectural configurations mentioned above). Note that the base architecture appears 8 times (1, on the number of neurons per layer, 1 on the number of layers, 1 on the number of labels and the 5 randomizations of weight initializations).

All experiments were executed in a machine with 2 NVIDIA V100 of 32GB, 2 Intel(R) Xeon(R) Platinum 8176 CPU @ 2.10GHz, and of 1.5TB RAM, for a total of around 3 days.

The code and results are fully open source\footnote{\url{https://github.com/asier-gutierrez/nn-similarity}} under MIT license.

\section{Results \& Discussion}
\label{sec:results}


\begin{table}
\centering
\begin{tabular}{@{}rlr@{}}
\toprule
\multicolumn{1}{l}{Number} & Experiment & \multicolumn{1}{l}{Index} \\ \midrule
1 & Layer size & 1-4 \\
2 & Number of layers & 5-9 \\
3 & Input order & 10-14 \\
4 & Number of labels & 15-19 \\ \bottomrule
\end{tabular}
\caption{Indices of the experiments of the distance matrices.}
\label{tab:experiment_indexing}
\end{table}

\begin{figure}
\centering
\begin{subfigure}{.48\textwidth}
  \centering
  \includegraphics[width=.9\linewidth]{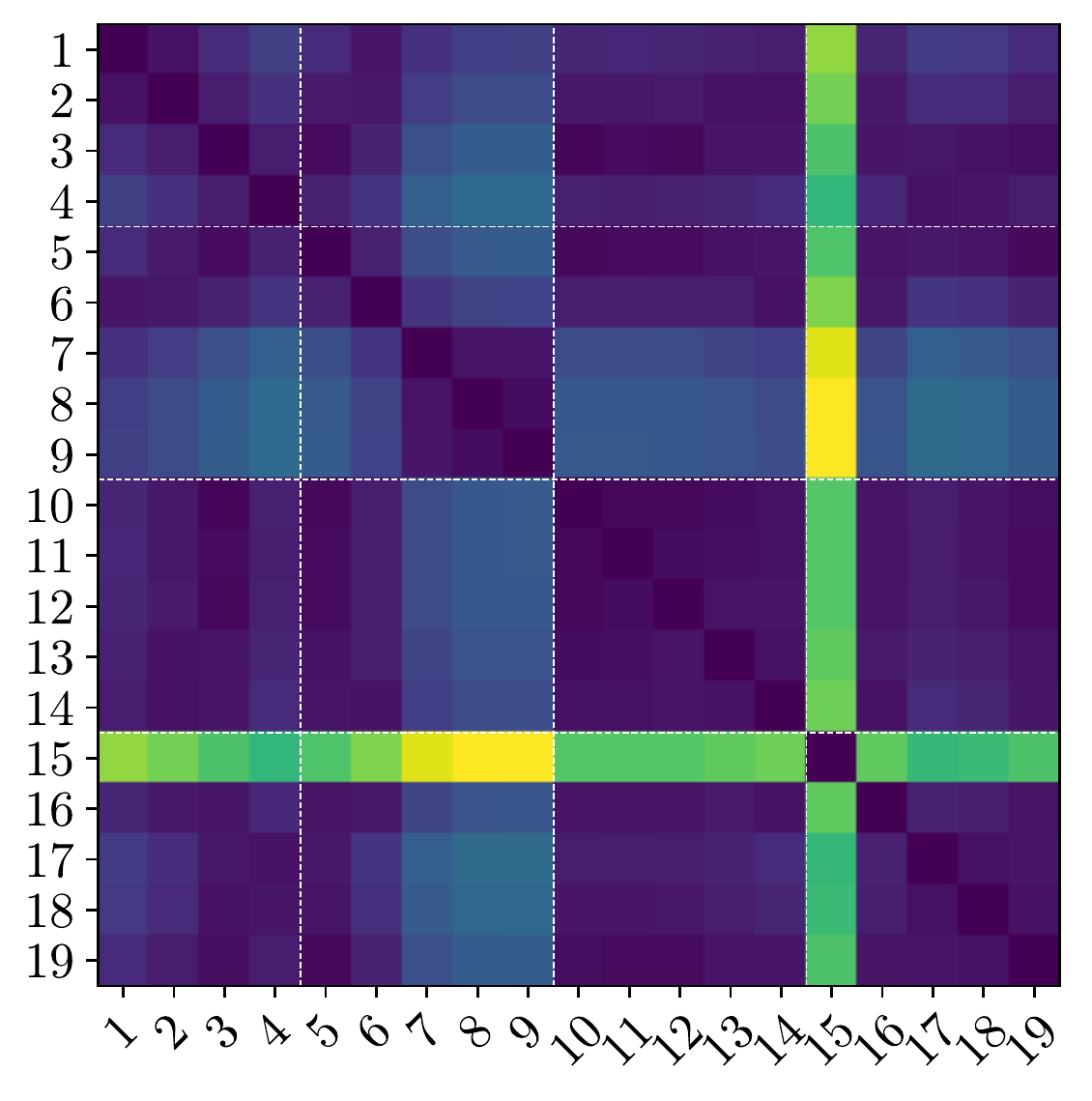}
  \caption{Reuters}
  \label{fig:control_experiments_reuters}
\end{subfigure}%
\begin{subfigure}{.48\textwidth}
  \centering
  \includegraphics[width=.9\linewidth]{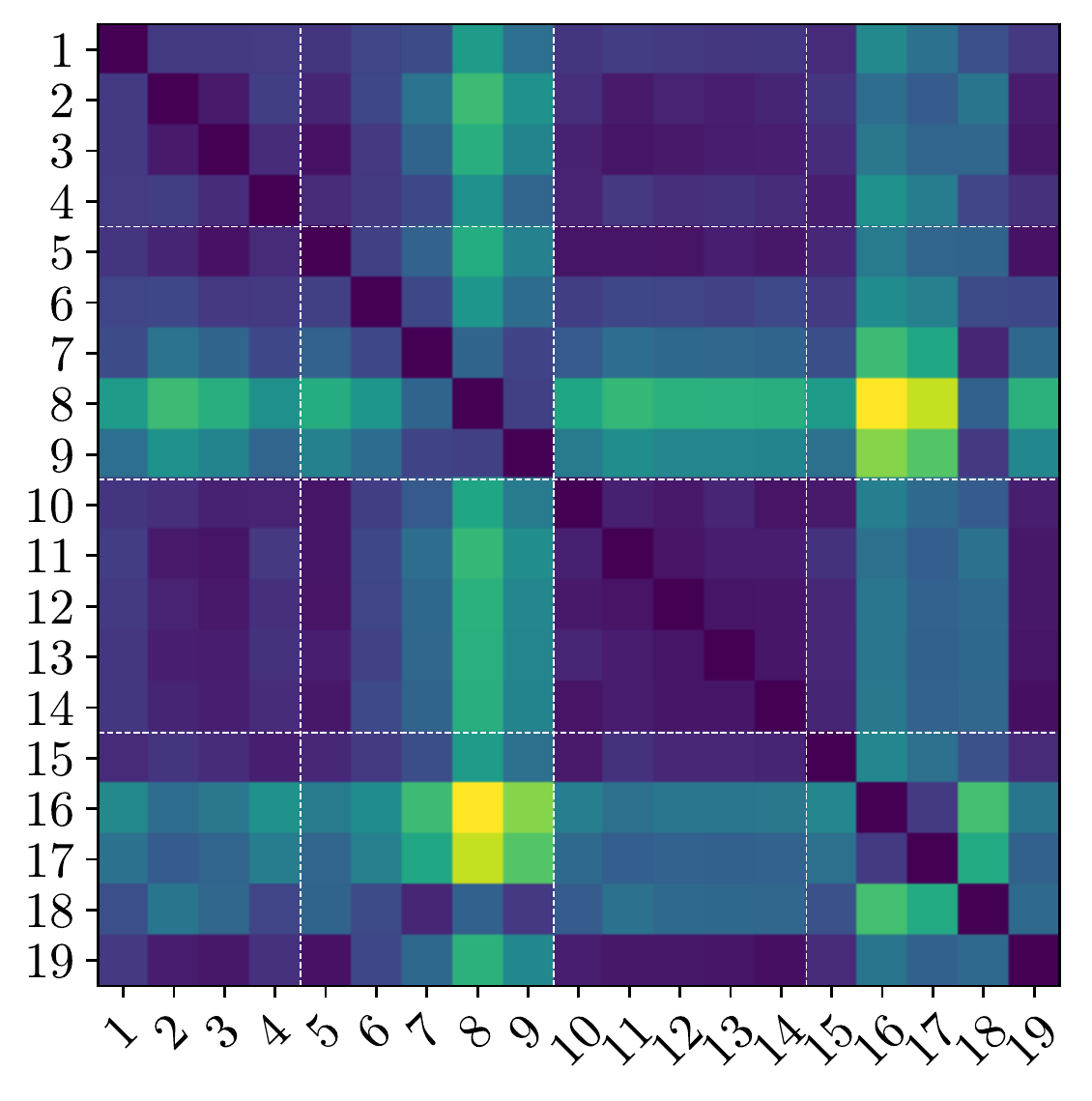}
  \caption{Language Identification}
  \label{fig:control_experiments_language_identification}
\end{subfigure}
\includegraphics[width=4.5cm]{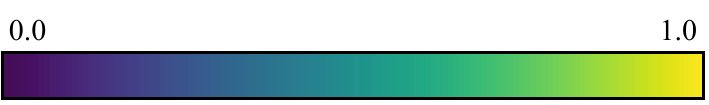}
\caption{Distance matrices using Silhouette discretization.}
\label{fig:control_experiments}
\end{figure}

\begin{figure}
\centering
\begin{subfigure}{.48\textwidth}
  \centering
  \includegraphics[width=.9\linewidth]{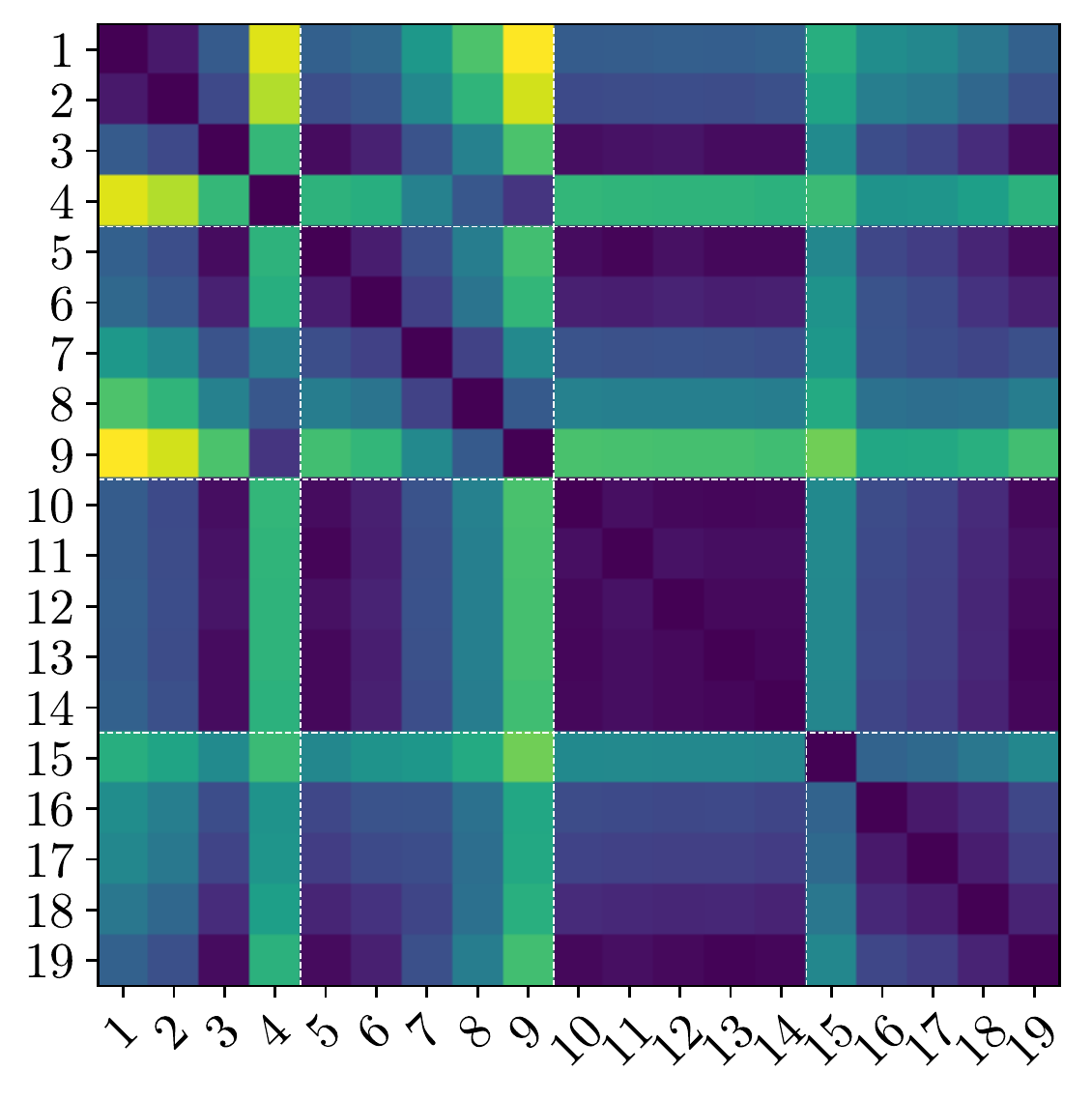}
  \caption{MNIST}
  \label{fig:fcn_cnn_relation_mnist}
\end{subfigure}%
\begin{subfigure}{.48\textwidth}
  \centering
  \includegraphics[width=.9\linewidth]{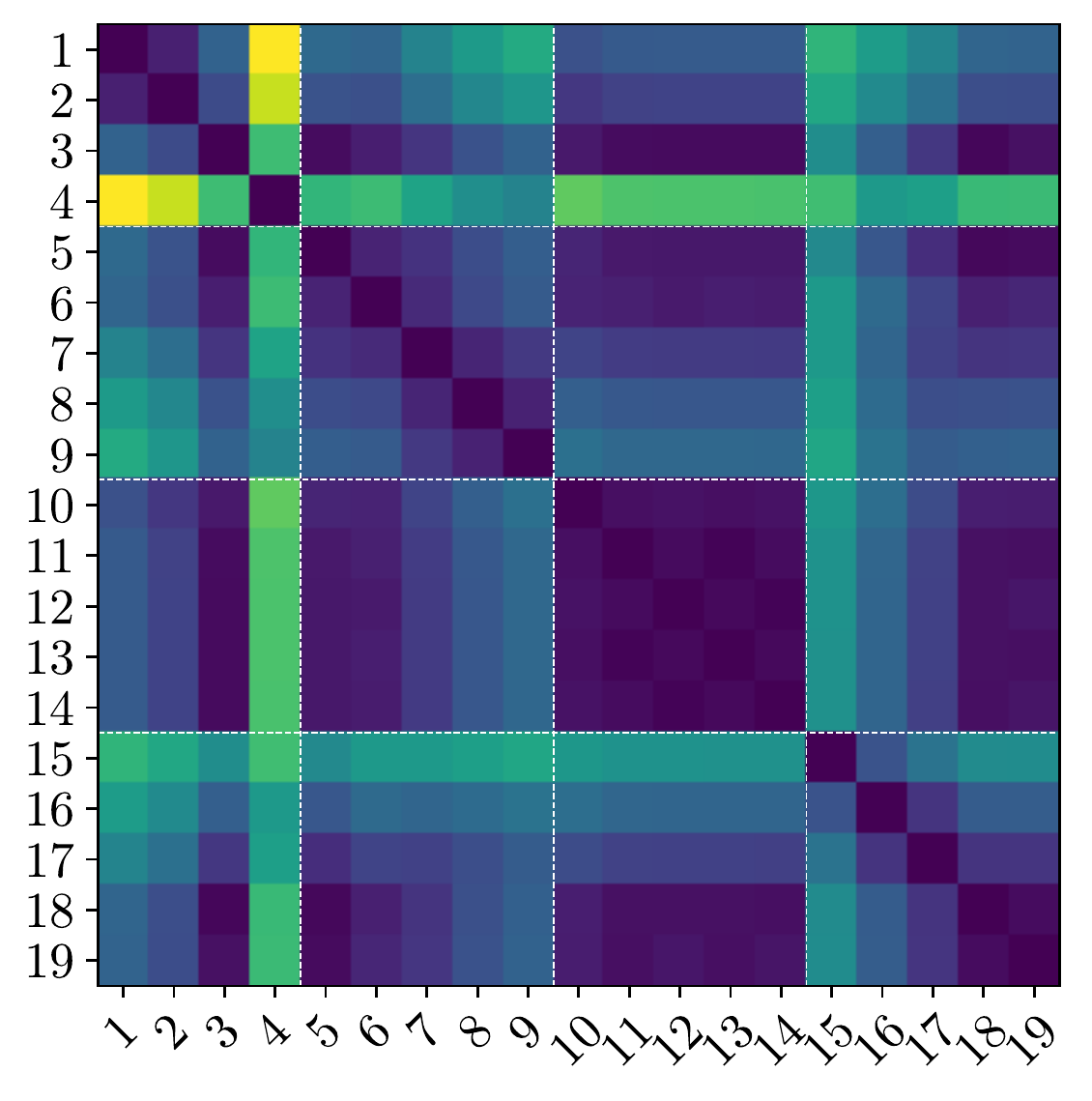}
  \caption{Fashion MNIST}
  \label{fig:fcn_cnn_relation_fashion_mnist}
\end{subfigure}
\begin{subfigure}{.48\textwidth}
  \centering
  \includegraphics[width=.9\linewidth]{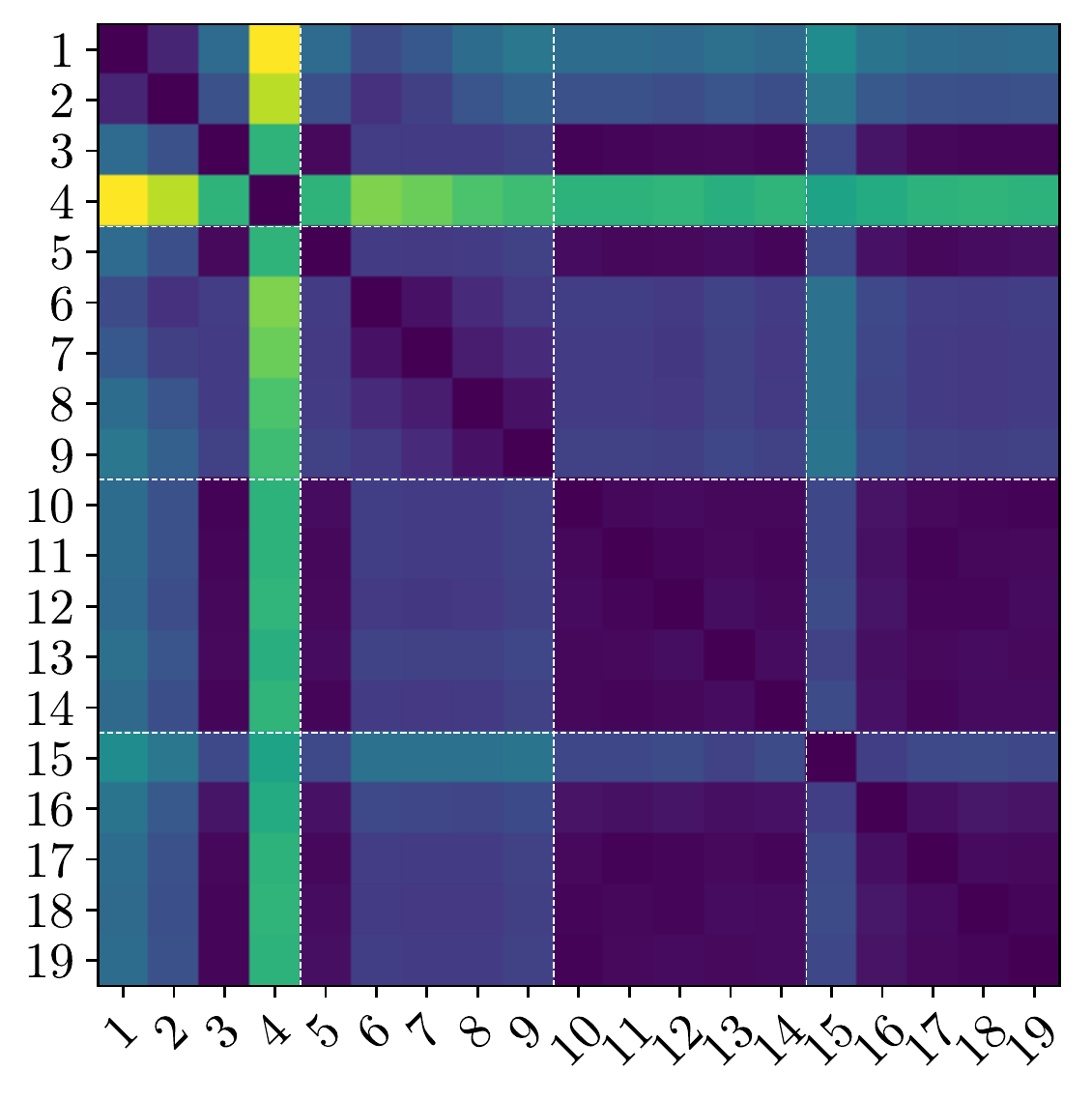}
  \caption{CIFAR-10}
  \label{fig:fcn_cnn_relation_cifar}
\end{subfigure}

\includegraphics[width=4.5cm]{img/general/temp_bar.pdf}
\caption{Distance matrices using Heat discretization.}
\label{fig:fcn_cnn_relation}
\end{figure}

Results from control experiments can be seen in the third group on Figures \ref{fig:control_experiments} and \ref{fig:fcn_cnn_relation}. In these figures, groups are separated visually using white dashed lines. Experiments groups are specified in Table \ref{tab:experiment_indexing}. Control experiments in all the images appear very dimmed, which means that they are very similar, as expected. Recall that the control experiments consist of $5$ (randomizations) $\times$ $5$ (executions) and that 25 different neural networks have been trained; each one of the network has more than 690,000 parameters that have been randomly initialized. After the training, results show that these networks have very close topological distance, as expected.

\begin{figure}
\centering
\begin{subfigure}{.48\textwidth}
  \centering
  \includegraphics[width=.9\linewidth]{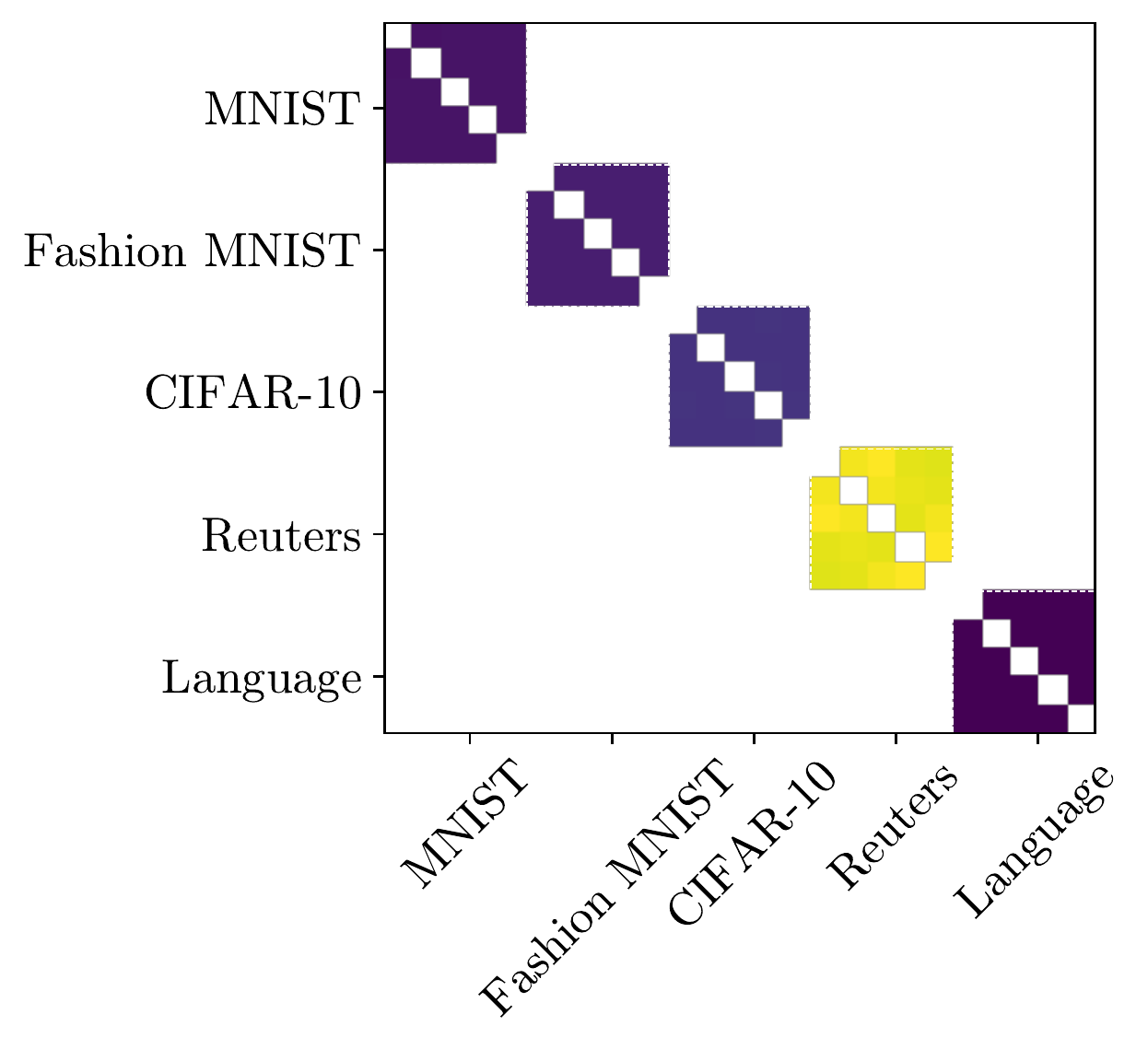}
  \caption{1-norm}
  \label{fig:control_experiments_norm_1}
\end{subfigure}%
\begin{subfigure}{.48\textwidth}
  \centering
  \includegraphics[width=.9\linewidth]{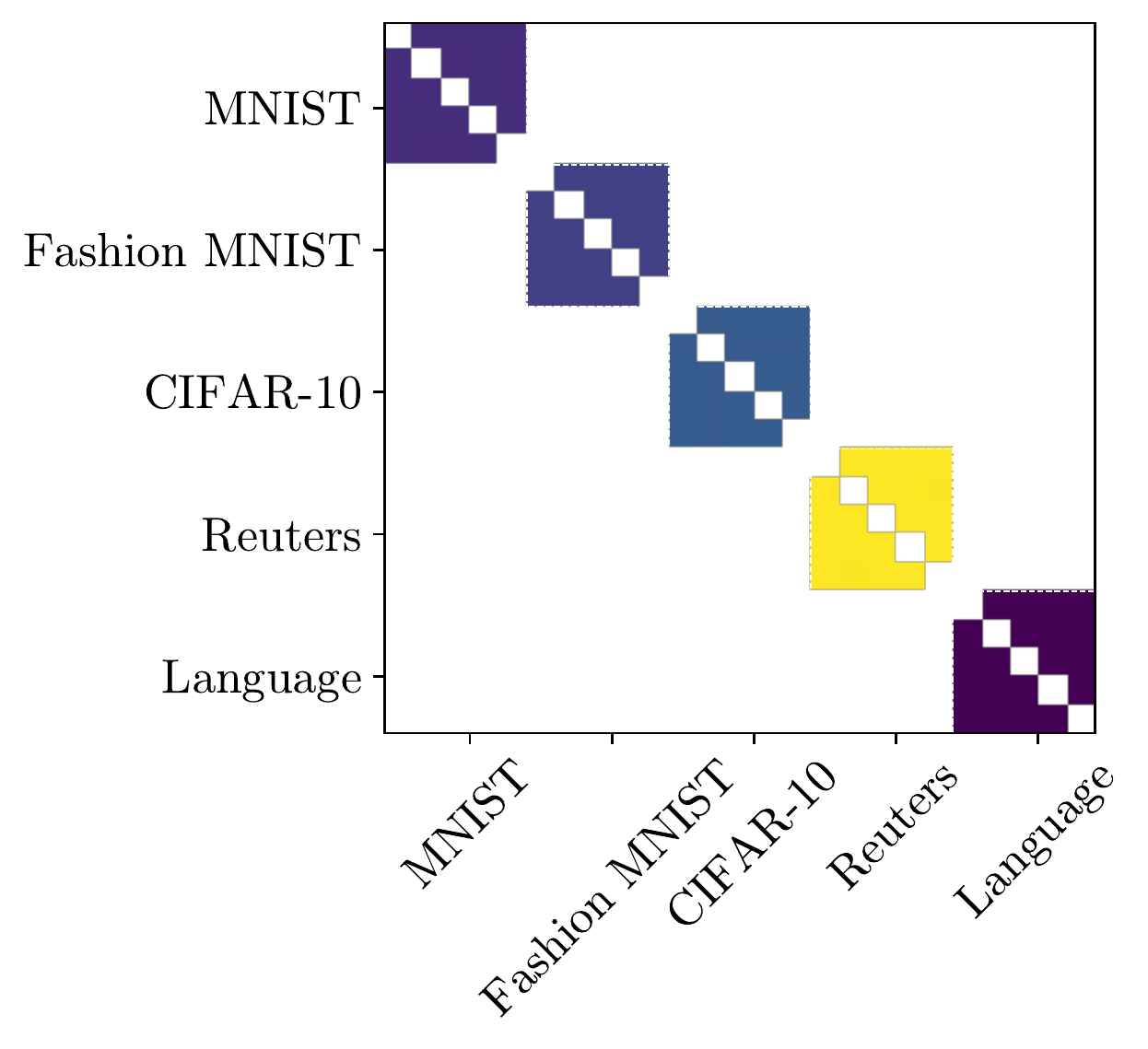}
  \caption{Frobenius norm}
  \label{fig:control_experiments_norm_frobenius}
\end{subfigure}
\includegraphics[width=4.5cm]{img/general/temp_bar.pdf}
\caption{Control experiments using norms.}
\label{fig:control_experiments_norms}
\end{figure}

\begin{table}
\centering
\begin{tabular}{@{}lrrrr@{}}
\toprule
Norm & \multicolumn{1}{l}{Minimum} & \multicolumn{1}{l}{Maximum} & \multicolumn{1}{l}{Mean} & \multicolumn{1}{l}{Standard deviation} \\ \midrule
1-Norm & 0.6683 & 4.9159 & 1.9733 & 1.5693 \\
Frobenius & 0.0670 & 0.9886 & 0.4514 & 0.3074 \\
\bottomrule
\end{tabular}
\caption{Normalized difference comparison of self-norm against the maximum mean distance of the experiment.}
\label{tab:norm-self}
\end{table}

For Figure \ref{fig:control_experiments_norms} we computed both 1-norm and Frobenius norm (the baselines) for graphs' adjacency matrices of control experiments. Note that as we ran the experiment five times, we make the mean for each value of the matrix. In order to show whether the resulting values are positive or negative, we subtract to the maximum difference of each dataset the norm of each cell separately, we take the absolute value and we divide by the maximum difference of each dataset. Therefore, we obtain five values per dataset. Table \ref{tab:norm-self} shows the statistics reflecting that the distance among the experiments are large and, thus, they are not characterizing any similarity but rather an important dissimilarity.

\begin{figure}
\centering
\includegraphics[scale=0.45]{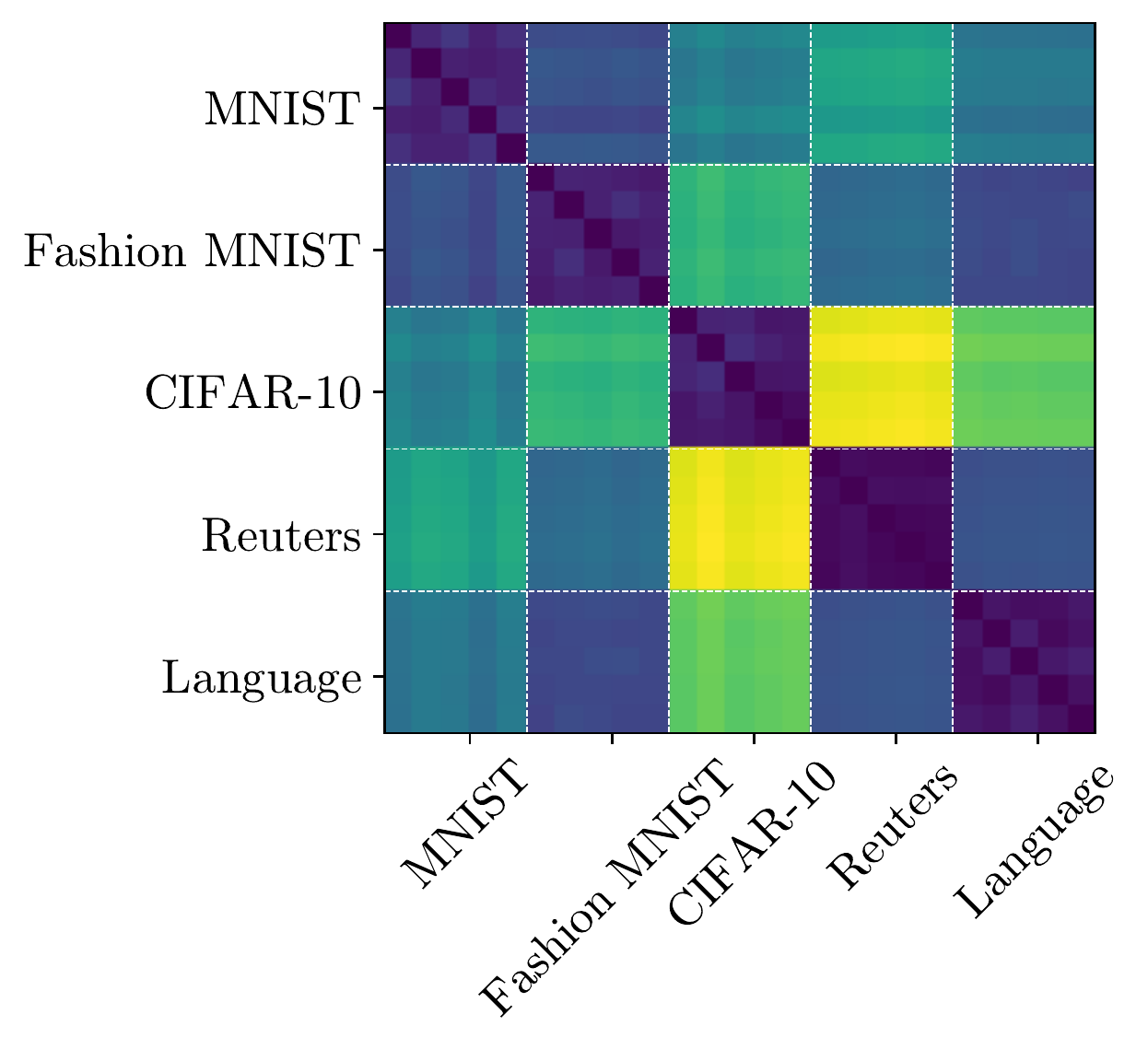}
\includegraphics[height=3.0cm, trim={0cm 0cm 0cm 3.4cm, clip}]{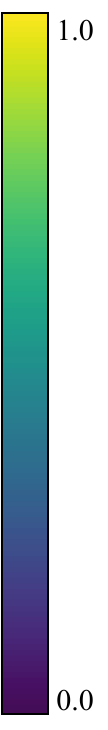}
\caption{Control experiment comparison matrix using Silhouette discretization.}
\label{fig:control_experiment_comparison_silhouette}
\end{figure}

In contrast, Figure \ref{fig:control_experiment_comparison_silhouette}, with our method (Silhouette), shows perfect diagonal of similarity blocks. In the corresponding numeric results, we obtained show small distances, as shown in Table \ref{tab:ph_dataset_distances}. We can appreciate that each dataset has its own hub. This confirms the validity of our proposed similarity measure.

The method we present also seems to capture some parts of hyperparameter setup. For instance, in Figure \ref{fig:fcn_cnn_relation} we can observe gradual increase of distances in the first group regarding layer size meaning that, as layer size increases, the topological distance increases too. Similarly, for the number of layers (second group) and number of labels (fourth group) the same situation holds. Note that in Fashion MNIST and CIFAR-10, the distances are dimmer because we are not dealing with the weights of the CNNs. Recall that the CNN acts as a frozen extractor and are pretrained for all runs (with the same weights), such that the MLP layers themselves are the only potential source of dissimilarity between runs.

\begin{table}
\centering
\begin{tabular}{@{}lrrrr@{}}
\cmidrule(l){2-5}
 & \multicolumn{2}{c}{Heat distance} & \multicolumn{2}{c}{Silhouette distance} \\ \midrule
Dataset & \multicolumn{1}{l}{Mean} & \multicolumn{1}{l}{Deviation} & \multicolumn{1}{l}{Mean} & \multicolumn{1}{l}{Deviation} \\ \midrule
MNIST & 0.0291 & 0.0100 & 0.1115 & 0.0364 \\
F. MNIST & 0.0308 & 0.0132 & 0.0824 & 0.0353 \\
CIFAR-10 & 0.0243 & 0.0068 & 0.0769 & 0.0204 \\
Language I. & 0.0159 & 0.0040 & 0.0699 & 0.0159 \\
Reuters & 0.0166 & 0.0051 & 0.0387 & 0.0112 \\ \bottomrule
\end{tabular}
\caption{PH distances across input order (control) experiments, normalized by dataset.}
\label{tab:ph_dataset_distances}
\end{table}

\clearpage
Thus, our characterization is sensitive to the architecture (e.g., if we increase the capacity, distances vary), but at the same time, as we saw before, it is not dataset-agnostic, meaning that it also captures whether two neural networks are learning the same problem or not.

In Figure \ref{fig:fcn_cnn_relation}, Fashion MNIST (Figure \ref{fig:fcn_cnn_relation_fashion_mnist}) and CIFAR (Figure \ref{fig:fcn_cnn_relation_cifar}) dataset results are interestingly different from those of MNIST (Figure \ref{fig:fcn_cnn_relation_mnist}) dataset. This is, presumably, because both Fashion MNIST and CIFAR use a pretrained CNN for the problem. Thus, we must analyze the results taking into account this perspective. The first fully connected layer size is important as it can avoid a bottleneck from the previous CNN output. Some works in the literature show that adding multiple fully connected layers does not necessarily enhance the prediction capability of CNNs \cite{DBLP:journals/corr/abs-1902-02771}, which is congruent with our results when adding fully connected layers (experiments 5 to 9) that result in dimmer matrices than the one from. Concerning the experiments on input order, there is slightly more homogeneity than in MNIST, again showing that the order of sample has negligible influence. Moreover, there could have been even more homogeneity taking into account that the fully connected network reduced its variance thanks to the frozen weights of the CNN. This also supports the fact that the CNN is the main feature extractor of the network. As in MNIST results, CIFAR results show that the topological properties are, indeed, a mapping of the practical properties of neural networks.

\begin{wrapfigure}[16]{L}{0.5\textwidth}
\centering
\includegraphics[scale=0.45]{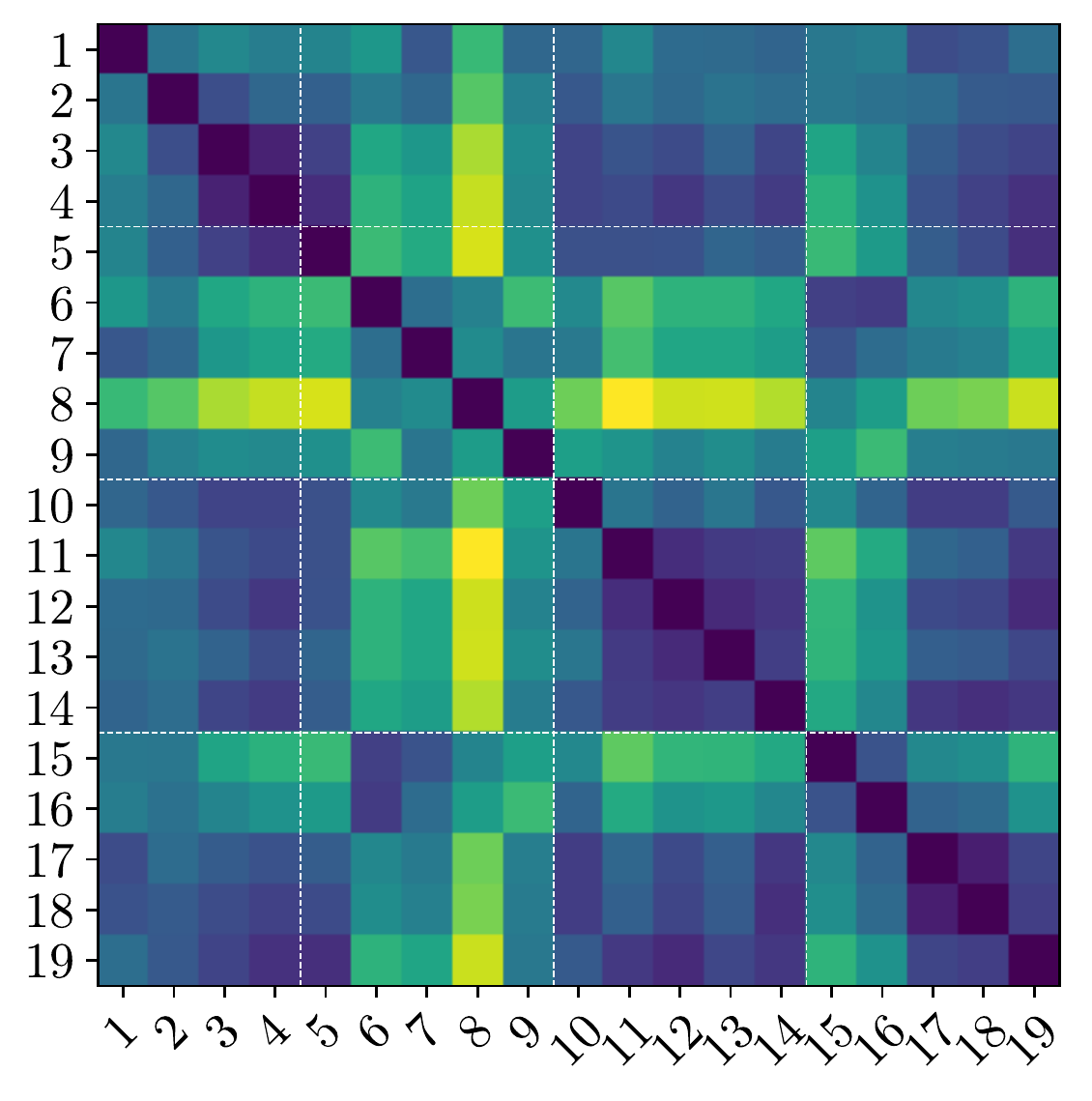}
\label{fig:landscape}
\includegraphics[height=3.0cm, trim={0cm 0cm 0cm 3.4cm, clip}]{img/general/vertical_temp_bar.pdf}
\caption{Language Identification dataset PH Landscape distance matrix.}
\end{wrapfigure}

We refer to the Supplementary Material for all distance matrices for all datasets and all distances, as well as for the standard deviations matrices and experiment group statistics.

\section{Conclusions \& Future Work}
\label{sec:conclusions}
Results from different experiments, in five different datasets from computer vision and natural language, lead to similar topological properties and are trivially interpretable, which yields to general applicability.

The bests discretizations chosen for this work are the Heat and Silhouette. They show better separation of experiment groups, and are effectively reflecting changes in a sensitive way. We also explored the Landscape discretization but it offers a very low interpretability and clearance. In other words, it is not helpful for comparing PH diagrams associated to neural networks. 

The most remarkable conclusion comes from the control experiments. The corresponding neural networks, with different input order but the same architecture, are very close to each other. The PH framework does, indeed, abstract away the specific weight values, and captures latent information from the networks, allowing comparisons to be based on the function they approximate. The selected neural network representation is reliable and complete, and yields coherent and meaningful results. Instead, the baseline measures, the 1-Norm and the Frobenius norm, implied an important dissimilarity between the experiments in the control experiments, meaning that they did not capture the fact that these neural networks were very similar in terms of the solved problem.

We conclude that our proposed characterization, does, indeed, capture meaningful information from neural network, and the computed distances can serve as an effective similarity measure between networks. To the best of our knowledge, this similarity measure between neural networks is the first of its kind.

As future work, we suggest adapting the method to different deep learning libraries and make it support popular neural architectures such as CNNs, Recurrent Neural Networks, and Transformers \cite{DBLP:journals/corr/VaswaniSPUJGKP17}. Finally, we suggest performing more analysis regarding the learning of a neural network, and trying to topologically answer the question of how a neural network learns.



\clearpage

\clearpage
\medskip
\bibliography{references}

\end{document}


\maketitle

\section*{Appendix I}
This Appendix contains mathematical definitions.

\begin{definition}[simplex]
 A \textit{k-simplex} is a k-dimensional polytope which is the convex hull of its k + 1 vertices. i.e. the set of all convex combinations $\lambda_0 v_0 + \lambda_1 v_1 + ... + \lambda_k v_k$ where $\lambda_0 + \lambda_1 + ... + \lambda_k = 1$ and
$0 \leq \lambda_j \leq 1 \hspace{2mm} \forall j \in \{0, 1, ..., k\}$.
\end{definition}
Some examples of simplices are:
\begin{itemize}
\item 0-simplex is a point.
\item 1-simplex is a line segment.
\item 2-simplex is a triangle.
\item 3-simplex is a tetrahedron.
\end{itemize}

\begin{definition}[simplicial complex]
A \textit{simplicial complex} $\mathcal{K}$ is a set of simplices that satisfies the following conditions:
\begin{enumerate}
\item Every subset (or face) of a simplex in $\mathcal{K}$ also belongs to $\mathcal{K}$.
\item For any two simplices $\sigma_1$ and $\sigma_2$ in $\mathcal{K}$, if $\sigma_1 \cap \sigma_2 \neq \emptyset$, then $\sigma_1 \cap \sigma_2$ is a common subset, or face, of both $\sigma_1$ and $\sigma_2$.
\end{enumerate}
\end{definition}


\begin{definition}[directed flag complex]
Let $G=(V,E)$ be a directed graph. The \textit{directed flag complex}  $FC(G)$ is defined to be the ordered simplicial complex whose $k$-simplices are all ordered $(k+1)$-cliques, i.e., $(k+1)$-tuples $\sigma=(v_0,v_1,\ldots,v_k)$, such that $v_i \in V \hspace{3mm} \forall i$, and $(v_i,v_j)\in E$ for $i<j$.
\end{definition}

We define the boundary, $\partial$, as a function that maps i-simplex to the sum of its (i-1)-dimensional faces. Formally speaking, for an i-simplex $\sigma=[v_0,\ldots, v_i]$, its \textit{boundary} $(\partial)$ is:
\begin{equation}
\partial_i\sigma = \sum^i_{j=0}[v_0,\ldots,\hat{v}_j,\ldots,v_i]
\end{equation}
where the hat indicates the $v_j$ is omitted.

We can expand this definition to $i$-chains. For an $i$-chain $c = c_i \sigma_i$, $\partial_i(c) = \sum_i c_i \partial_i \sigma_i$.

We can now distinguish two special types of chains using the boundary map that will be useful to define homology:
\begin{itemize}
\item The first one is an \textit{$i$-cycle}, which is defined as an $i$-chain with empty boundary. In other words, an $i$-chain $c$ is an $i$-cycle if and only if $\partial_i(c) = 0$, i.e. $c \in Ker(\partial_i)$.
\item An $i$-chain $c$ is \textit{$i$-boundary} if there
exists an $(i + 1)$-chain $d$ such that $c = \partial_{i+1}(d)$, i.e. $c \in Im(\partial_{i+1})$.
\end{itemize}


\begin{definition}[graph]
A \textit{graph} $G$ is a pair $(V,E)$, where $V$ is a finite set referred to as the vertices or nodes of $G$, and $E$ is a subset of the set of unordered pairs $e=\{u,v\}$ of distinct points in $V$, which we call the edges of $G$. Geometrically the pair $\{u,v\}$ indicates that the vertices $u$ and $v$ are adjacent in $G$. A directed graph, or a digraph, is similarly a pair $(V,E)$ of vertices $V$ and edges $E$, except the edges are ordered pairs of distinct vertices, i.e.,the pair $(u,v)$ indicates that there is an edge from $u$ to $v$ in $G$. In a digraph, we allow reciprocal edges, i.e., both $(u,v)$ and $(v,u)$ may be edges in $G$, but we exclude loops, i.e., edges of the form $(v,v)$.
\end{definition}

\begin{definition}[homology group]
Given these two special subspaces, $i$-cycles $Z_i(K)$ and $i$-boundaries $B_i(K)$ of $C_i(K)$, we now take the quotient space of $B_i(K)$ as a subset of $Z_i(K)$. In this quotient space, there are only the $i$-cycles that do not bound an $(i+1)$-complex, or $i$-voids of $K$. This quotient space is called $i$-th homology group of the simplicial complex $K$:
\begin{equation}
H_i(K) = \frac{Z_i(K)}{B_i(K)} = \frac{Ker(\partial_i)}{Im(\partial_{i+1})}
\end{equation}
\end{definition}
where $Ker$ and $Im$ are the function kernel and image respectively.

The dimension of $i$-th homology is called the $i$-th \textit{Betti number} of $K$, $\beta_i(K)$, where:
\begin{equation}
\beta_i(K) = dim(Ker(\partial_i)) - dim(Im(\partial_{i+1})) 
\end{equation}

\begin{definition}[Wasserstein distance]
The \textit{$p$-Wasserstein distance} between two PDs $D_1$ and $D_2$ is the infimum over all bijections: $\gamma: D_1 \to D_2$ of:
\begin{equation}
d_{W}(D_1, D_2) = \Big(\sum_{x \in D_1} ||x - \gamma(x)||_\infty^p \Big)^{1/p}
\end{equation}
where $||-||_\infty$ is defined for $(x,y) \in \mathbb{R}^2$ by $\max\{|x|,|y|\}$.
The limit $p \to \infty$ defines the \textit{Bottleneck distance}. More explicitly, it is the infimum over the same set of bijections of the value
\begin{equation}
d_B(D_1, D_2) = \sup_{x \in D_1} ||x - \gamma(x)||_{\infty}.
\end{equation}
\end{definition}

\begin{definition}[Persistence landscape]
Given a collection of intervals $\{(b_i, d_i)\}_{i \in I}$ that compose a PD, its \emph{persistence landscape} is the set of functions $\lambda_k: \mathbb R \to \overline{\mathbb R}$ defined by letting $\lambda_k(t)$ be the $k$-th largest value of the set $\{\Lambda_i(t)\}_ {i \in I}$ where:
\begin{equation}
\Lambda_i(t) = \left[ \min \{t-b_i, d_i-t\}\right]_+    
\end{equation}
and $c_+ := \max(c,0)$. The function $\lambda_k$ is referred to as the $k$-layer of the persistence landscape.

Now we define a vectorization of the set of real-valued function that compose PDs on $\mathbb N \times \mathbb R$. For any $p = 1,\dots,\infty$ we can restrict attention to PDs $D$ whose associated persistence landscape $\lambda$ is $p$-integrable, that is to say,
\begin{equation}
    \label{equation:persistence_landscape_norm}
    ||\lambda||_p = \left( \sum_{i \in \mathbb N} ||\lambda_i||^p_p \right)^{1/p}
\end{equation}
is finite. In this case, we refer to Equation (\ref{equation:persistence_landscape_norm}) as the $p$-landscape norm of $D$. For $p = 2$, we define the value of the \emph{landscape kernel} or similarity of two vectorized PDs $D$ and $E$ as
\begin{equation}
    \langle \lambda, \mu \rangle = \left(\sum_{i \in \mathbb N} \int_{\mathbb R} |\lambda_i(x) - \mu_i(x)|^2\, dx\right)^{1/2}
\end{equation}
where $\lambda$ and $\mu$ are their associated persistence landscapes.
\end{definition}
$\lambda_k$ is geometrically described as follows. For each $i \in I$, we draw an isosceles triangle with base the interval $(b_i, d_i)$ on the horizontal $t$-axis, and sides with slope $1$ and $-1$. This subdivides the plane into a number of polygonal regions that we label by the number of triangles contained on it. If $P_k$ is the union of the polygonal regions with values at least $k$, then the graph of $\lambda_k$ is the upper contour of $P_k$, with $\lambda_k(a) = 0$ if the vertical line $t=a$ does not intersect $P_k$.

\begin{definition}[Weighted silhouette]
Let $D = \{(b_i, d_i)\}_{i \in I}$ be a PD and $w = \{w_i\}_{i \in I}$ a set of positive real numbers. The silhouette of $D$ weighted by $w$ is the function $\phi: \mathbb R \to \mathbb R$ defined by:
\begin{equation}
    \phi(t) = \frac{\sum_{i \in I}w_i \Lambda_i(t)}{\sum_{i \in I}w_i},
\end{equation}
where
\begin{equation}
    \Lambda_i(t) = \left[ \min \{t-b_i, d_i-t\}\right]_+
\end{equation}
and $c_+ := \max(c,0)$ When $w_i = \vert d_i - b_i \vert^p$ for $0 < p \leq \infty$ we refer to $\phi$ as the $p$-power-weighted silhouette of $D$. It defines a vectorization of the set of PDs on the vector space of continuous real-valued functions on $\mathbb R$.
\end{definition}

\begin{definition}[Heat vectorizations]
Considering PD as the support of Dirac deltas, one can construct, for any $t > 0$, two vectorizations of the set of PDs to the set of continuous real-valued function on the first quadrant $\mathbb{R}^2_{>0}$. The heat vectorization is constructed for every PD $D$ by solving the heat equation:

\begin{eqnarray}
 \label{equation:heat_equation}
    \begin{split}\begin{aligned}
    \Delta_x(u) &= \partial_t u && \text{on } \Omega \times \mathbb R_{>0} \\
    u &= 0 && \text{on } \{x_1 = x_2\} \times \mathbb R_{\geq 0} \\
    u &= \sum_{p \in D} \delta_p && \text{on } \Omega \times {0} \end{aligned}\end{split}
\end{eqnarray}
where $\Omega = \{(x_1, x_2) \in \mathbb R^2\ |\ x_1 \leq x_2\}$, then solving the same equation after precomposing the data of Equation (\ref{equation:heat_equation}) with the change of coordinates $(x_1, x_2) \mapsto (x_2, x_1)$, and defining the image of $D$ to be the difference between these two solutions at the chosen time $t$.

We recall that the solution to the heat equation with initial condition given by a Dirac delta supported at $p \in \mathbb R^2$ is:
\begin{equation}
    \frac{1}{4 \pi t} \exp\left(-\frac{||p-x||^2}{4t}\right)
\end{equation}
To highlight the connection with normally distributed random variables, it is customary to use the the change of variable $\sigma = \sqrt{2t}$.
\end{definition}

For a complete reference on vectorized persistence summaries and PH approximated metrics, see \citet{tauzin2020giottotda, Berry2020FunctionalSO} and Giotto-TDA package documentation appendix\footnote{\url{https://giotto-ai.github.io/gtda-docs/0.3.1/theory/glossary.html\#persistence-landscape}}.

\begin{figure}[h]
\centering
    \includegraphics[width=340px]{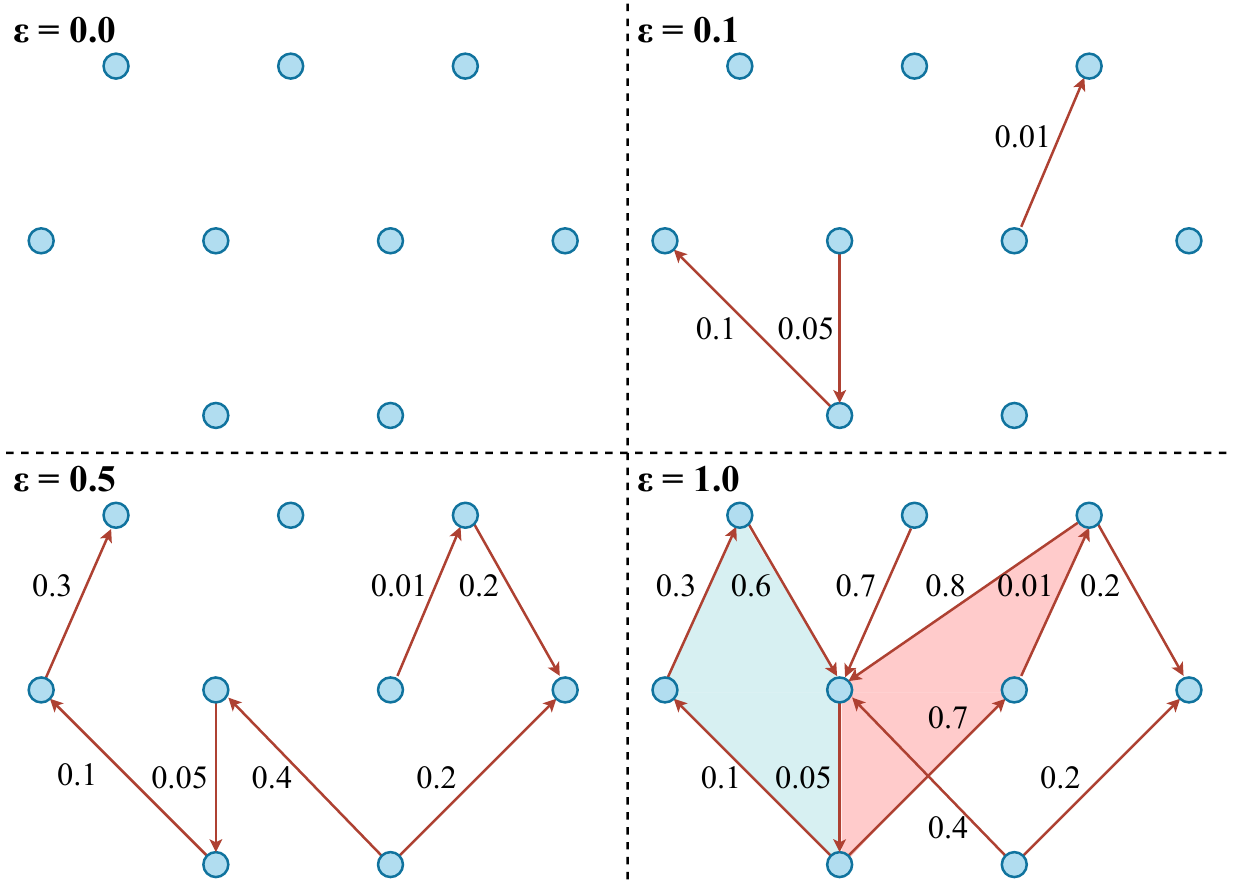}
    \caption{MLP Simplicial complex filtration example.} 
    \label{fig:simplicial_complex_filtration}
\end{figure}

Figure \ref{fig:simplicial_complex_filtration} shows a neural network filtration example. Note that most of the edges have been omitted for clarity and, for the same reason, $\epsilon$ evolution has been discretized.

\begin{figure}[!ht]
\centering
    \includegraphics[width=240px]{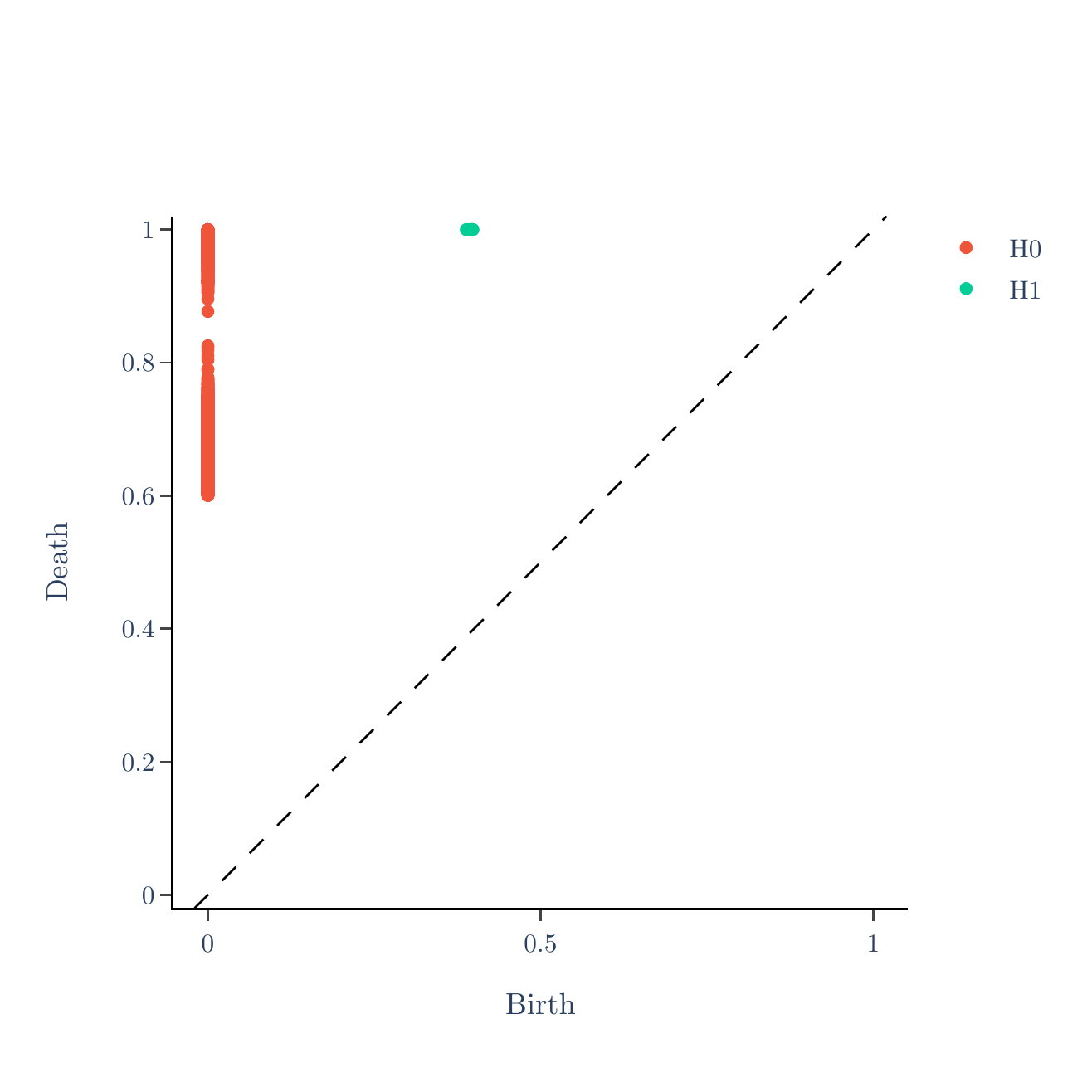}
    \caption{A Persistence Homology diagram.} 
    \label{fig:ph_diagram}
\end{figure}

\clearpage
\section*{Appendix II}

\begin{figure}[H]
\centering
\begin{tabular}{cc}
\subfloat[Norm-1 distance means.]{\includegraphics[width = 3in]{img/additional_material/control_experiments/1_distance.pdf}} &
\subfloat[Norm-1 distance std.]{\includegraphics[width = 3in]{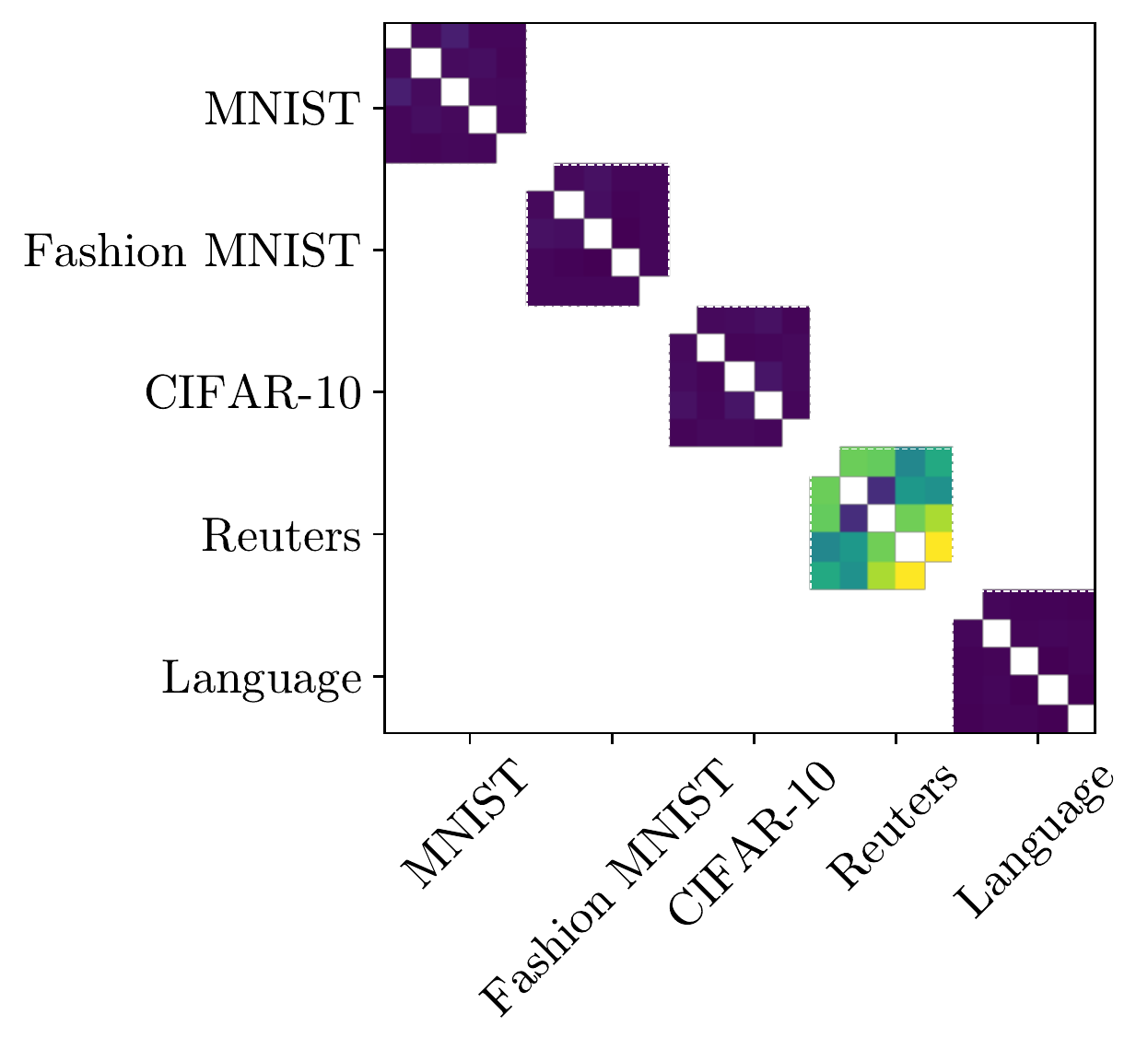}} \\
\subfloat[Frobenius norm distance means.]{\includegraphics[width = 3in]{img/additional_material/control_experiments/Frobenius_distance.pdf}} &
\subfloat[Frobenius norm distance std.]{\includegraphics[width = 3in]{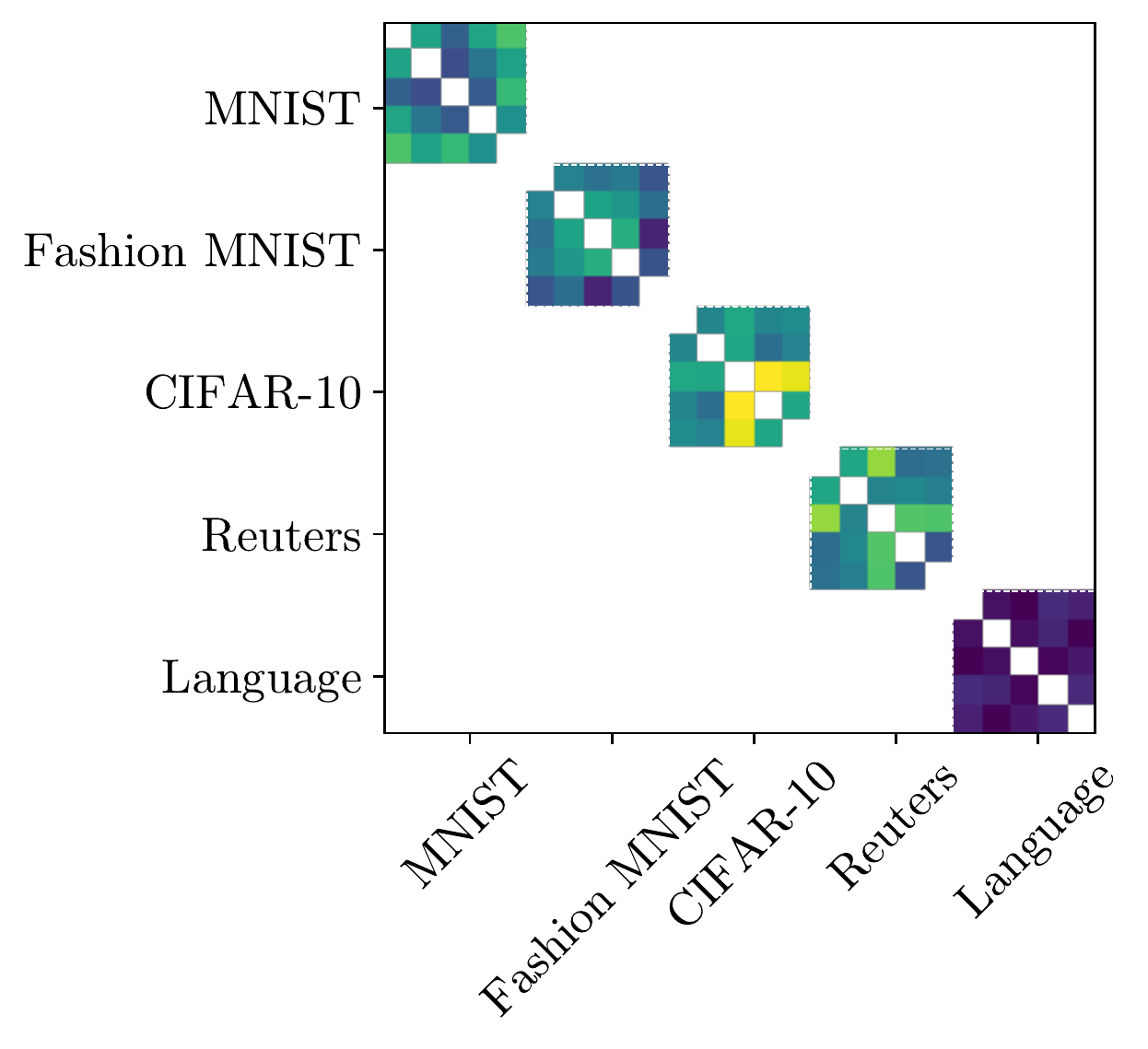}} \\
\multicolumn{2}{c}{\includegraphics[width = 5.5cm]{img/general/temp_bar.pdf}} \\
\end{tabular}
\caption{Norm distances among control experiments. 5 runs $\times$ 5 randomizations.}
\label{fig:norm-distances}
\end{figure}

\begin{table}[H]
\centering
\begin{tabular}{@{}lrrrr@{}}
\toprule
Norm & \multicolumn{1}{l}{Minimum} & \multicolumn{1}{l}{Maximum} & \multicolumn{1}{l}{Mean} & \multicolumn{1}{l}{Standard deviation} \\ \midrule
1-Norm & 0.6683 & 4.9159 & 1.9733 & 1.5693 \\
Frobenius & 0.0670 & 0.9886 & 0.4514 & 0.3074 \\
\bottomrule
\end{tabular}
\caption{Normalized difference comparison of self-norm against the maximum mean distance of the experiment.}
\label{tab:norm-self}
\end{table}

\begin{figure}[H]
\centering
\begin{tabular}{ccc}
\subfloat[Heat distance.]{\includegraphics[width = 2in]{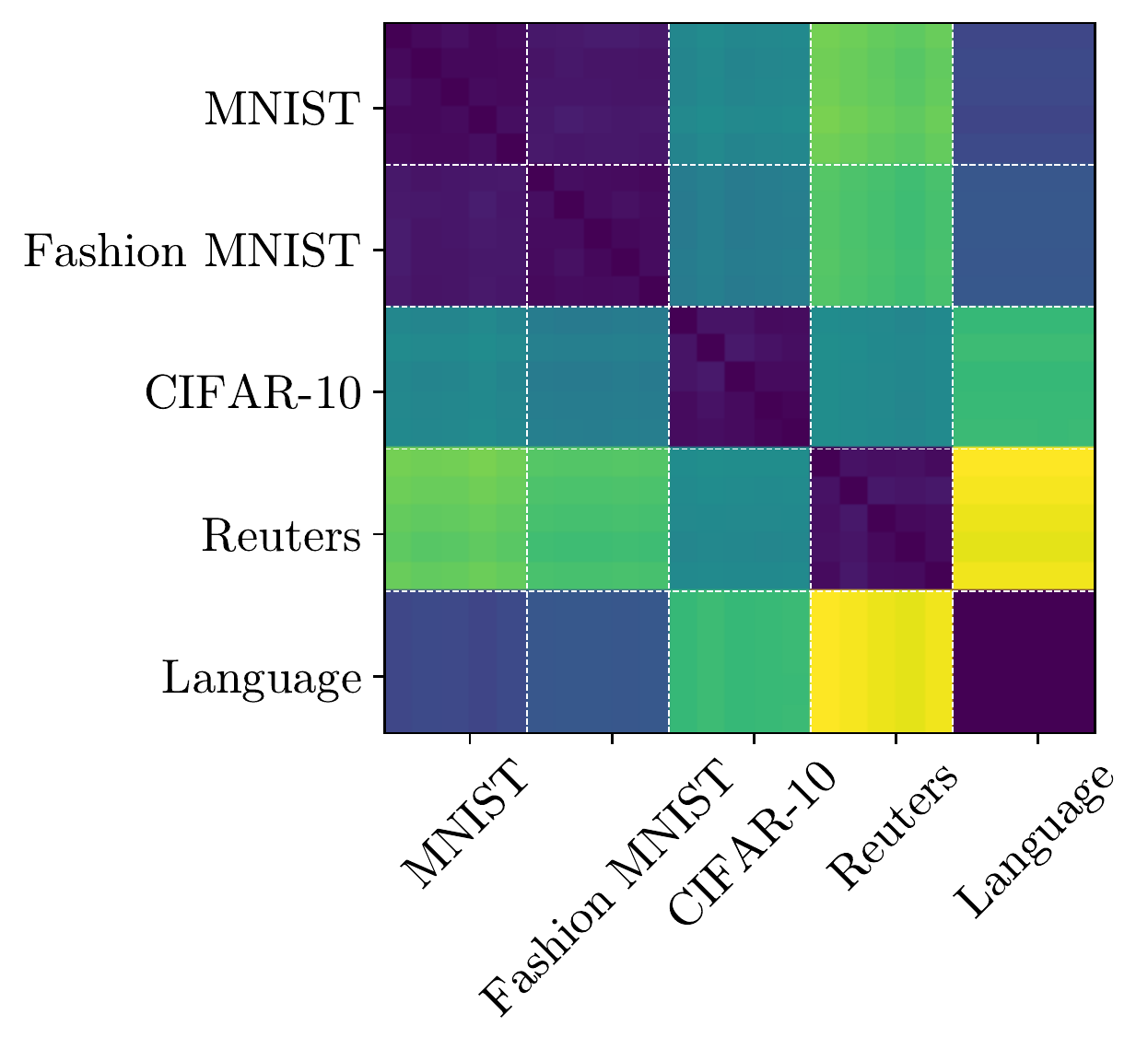}} &
\subfloat[Silhouette distance.]{\includegraphics[width = 2in]{img/additional_material/control_experiments/Silhouette_distance.pdf}} &
\subfloat[Landscape distance.]{\includegraphics[width = 2in]{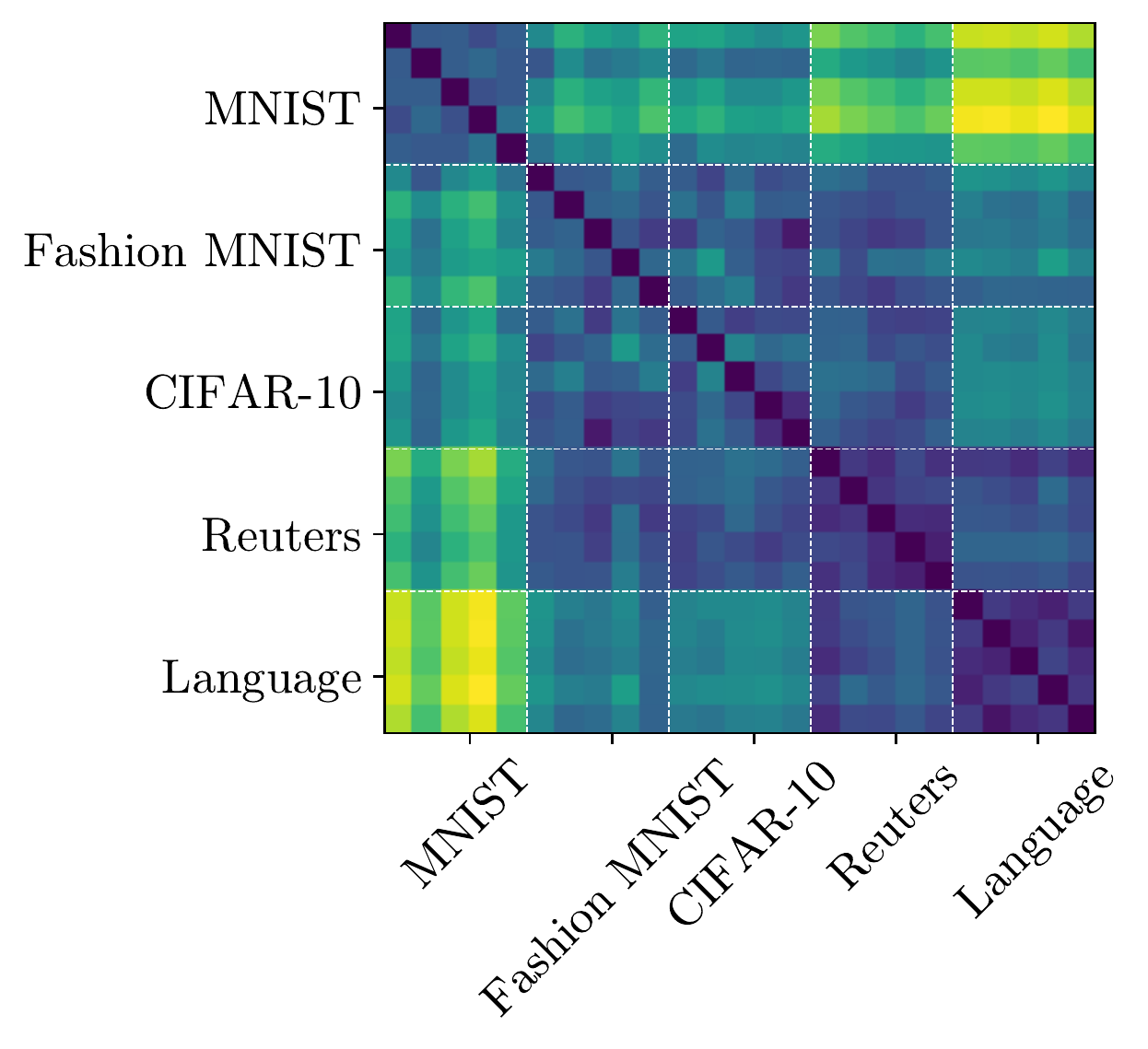}}
\\
\multicolumn{3}{c}{\includegraphics[width = 5.5cm]{img/general/temp_bar.pdf}} \\
\end{tabular}
\caption{Topological distance means of control experiments.}
\label{fig:ap_control_experiments_means}
\end{figure}

\begin{figure}[H]
\centering
\begin{tabular}{ccc}
\subfloat[Heat distance.]{\includegraphics[width = 2in]{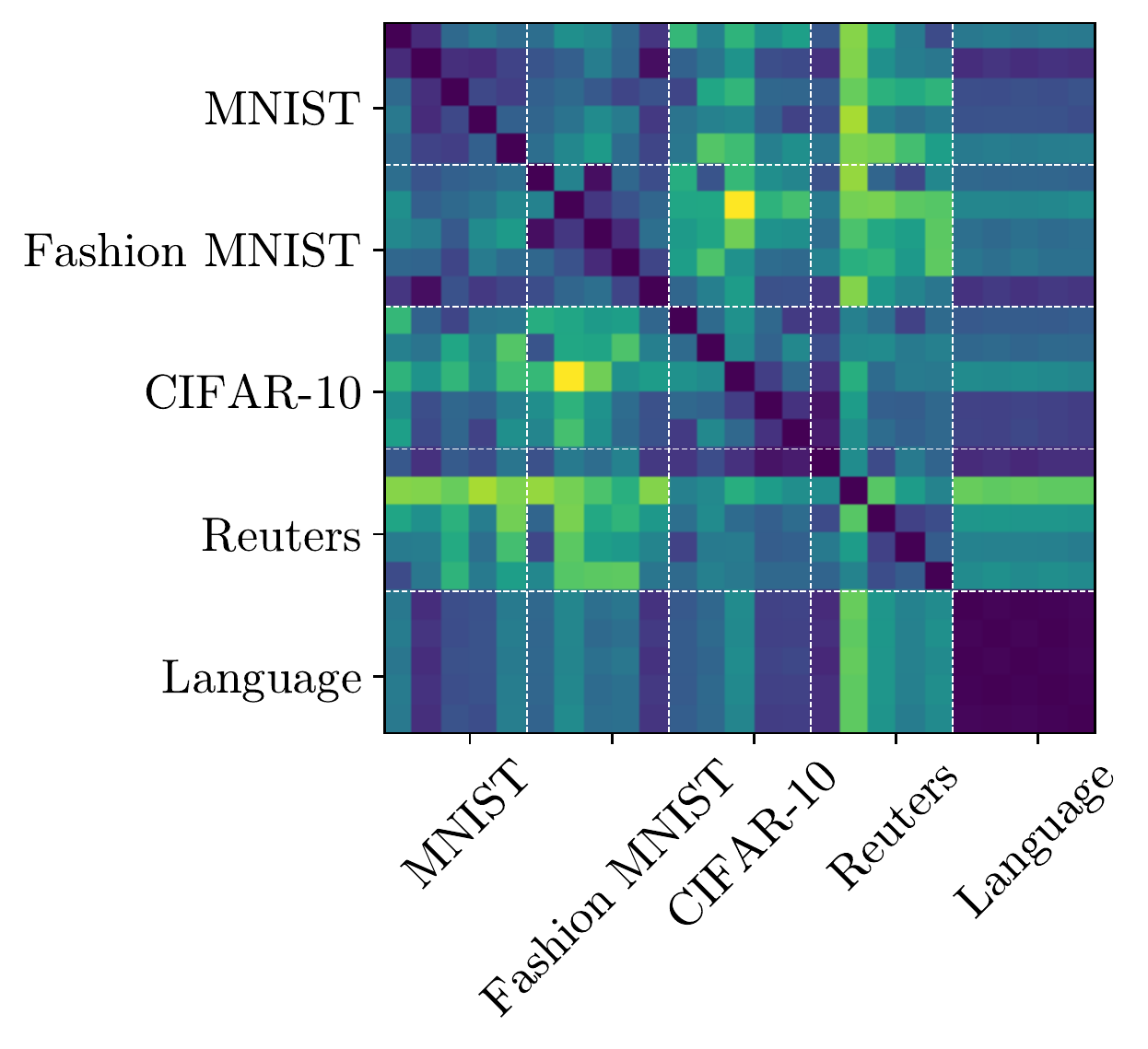}} &
\subfloat[Silhouette distance.]{\includegraphics[width = 2in]{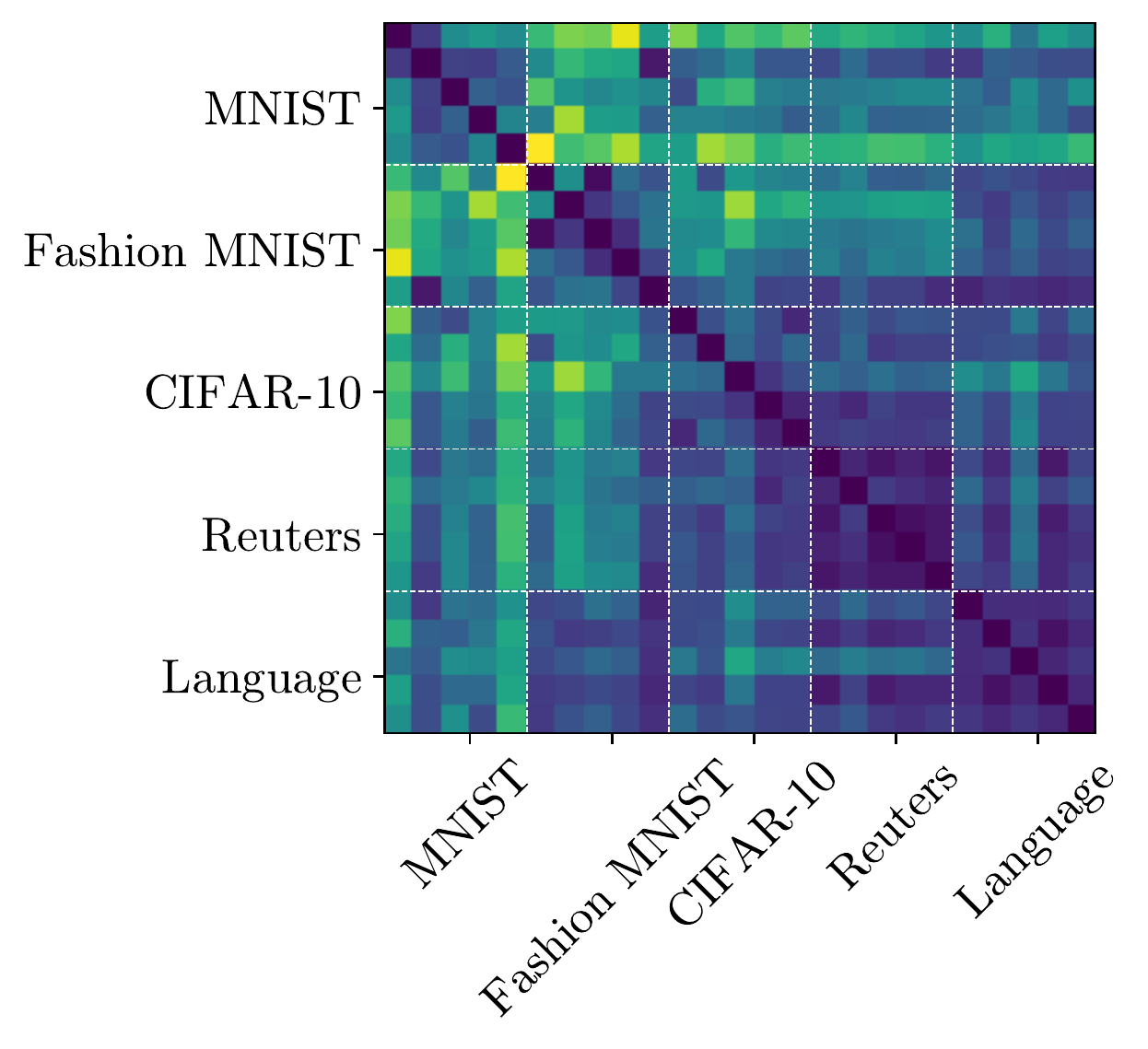}} &
\subfloat[Landscape distance.]{\includegraphics[width = 2in]{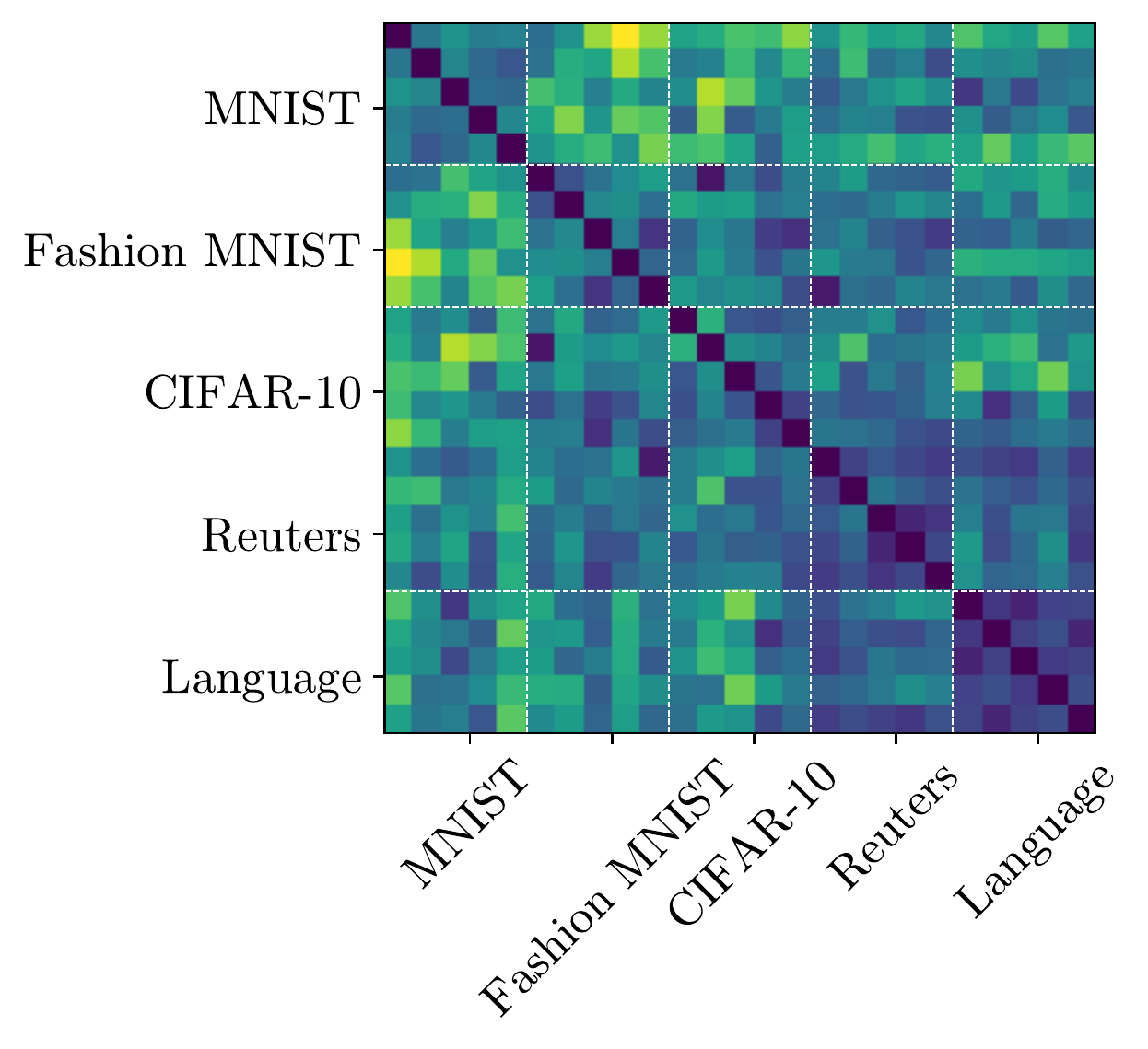}}
\\
\multicolumn{3}{c}{\includegraphics[width = 5.5cm]{img/general/temp_bar.pdf}} \\
\end{tabular}
\caption{Topological distance standard deviations of control experiments.}
\label{fig:ap_control_experiments_means}
\end{figure}

\clearpage
\section*{Appendix III}
This Appendix contains statistics about experiment groups by dataset.

\begin{table}[h]
\centering
\begin{tabular}{|r|l|r|r|}
\hline
\multicolumn{1}{|l|}{\textbf{Experiment group}} & \textbf{Experiment} & \multicolumn{1}{l|}{\textbf{Value}} & \multicolumn{1}{l|}{\textbf{Index}} \\ \hline
\multirow{4}{*}{1} & \multirow{4}{*}{Layer size} & 128 & 1 \\
 &  & 256 & 2 \\
 &  & 512 & 3 \\
 &  & 1024 & 4 \\ \hline
\multirow{5}{*}{2} & \multirow{5}{*}{Number of layers} & 2 & 5 \\
 &  & 4 & 6 \\
 &  & 6 & 7 \\
 &  & 8 & 8 \\
 &  & 10 & 9 \\ \hline
3 & Input order & NA & 10-14 \\ \hline
\multirow{5}{*}{4} & \multirow{5}{*}{Number of labels} & 2 & 15 \\
 &  & 4 (M, FM, C), 6 (R), 3 (L) & 16 \\
 &  & 6 (M, FM, C), 12 (R), 4 (L) & 17 \\
 &  & 8 (M, FM, C), 23 (R), 6 (L) & 18 \\
 &  & 10 (M, FM, C), 46 (R), 7 (L) & 19 \\ \hline
\end{tabular}
\caption{Indices of the experiments of the distance matrices. M is for MNIST, FM for Fashion MNIST, C for CIFAR-10, R for Reuters and L for Language Identification.}
\label{tab:experiment_indexing}
\end{table}

\begin{figure}[H]
\centering
\begin{tabular}{ccc}
\subfloat[Heat distance.]{\includegraphics[width = 2in]{img/results/mnist/heat_distance_matrices.pdf}} &
\subfloat[Silhouette distance.]{\includegraphics[width = 2in]{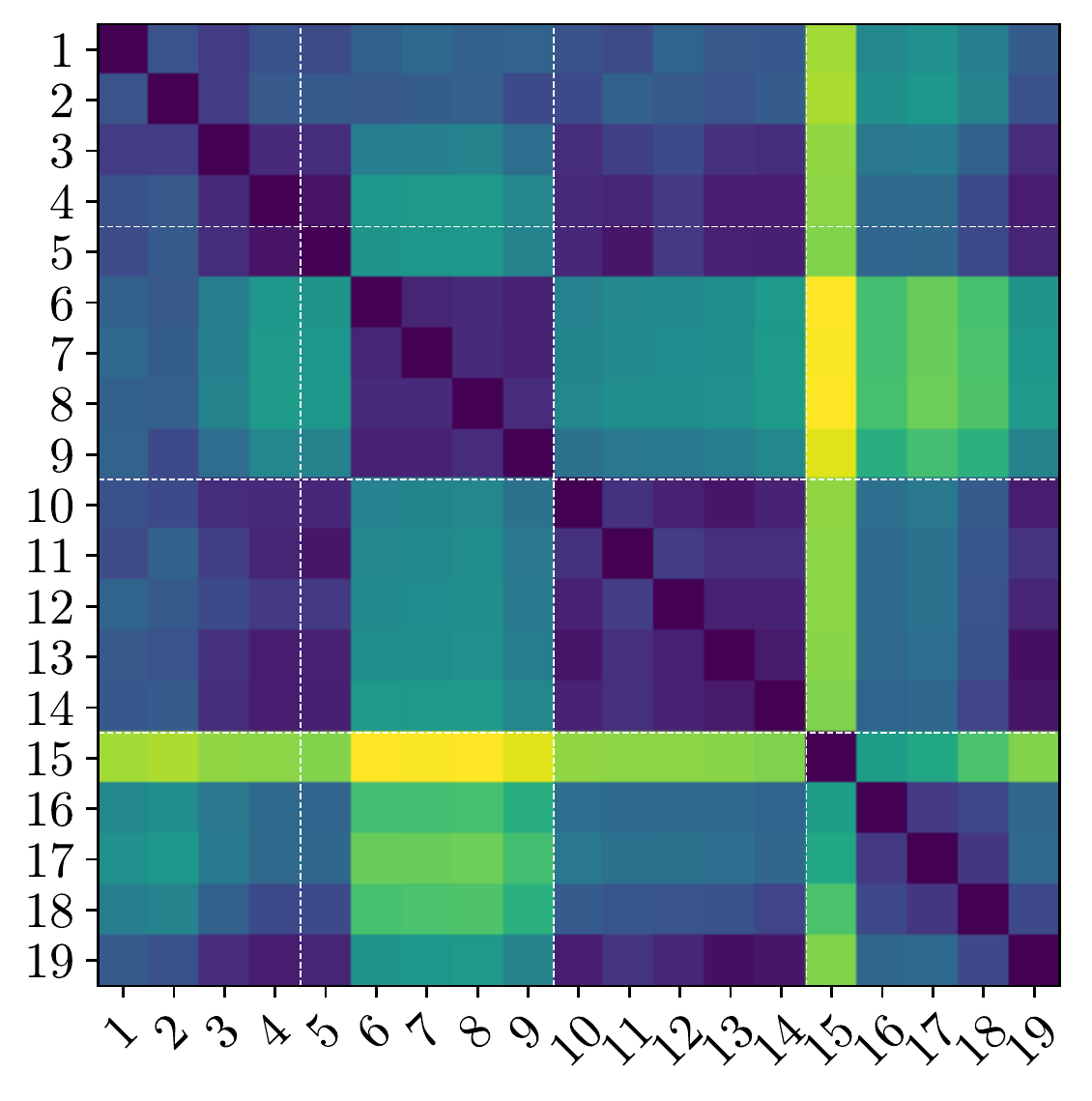}} &
\subfloat[Landscape distance.]{\includegraphics[width = 2in]{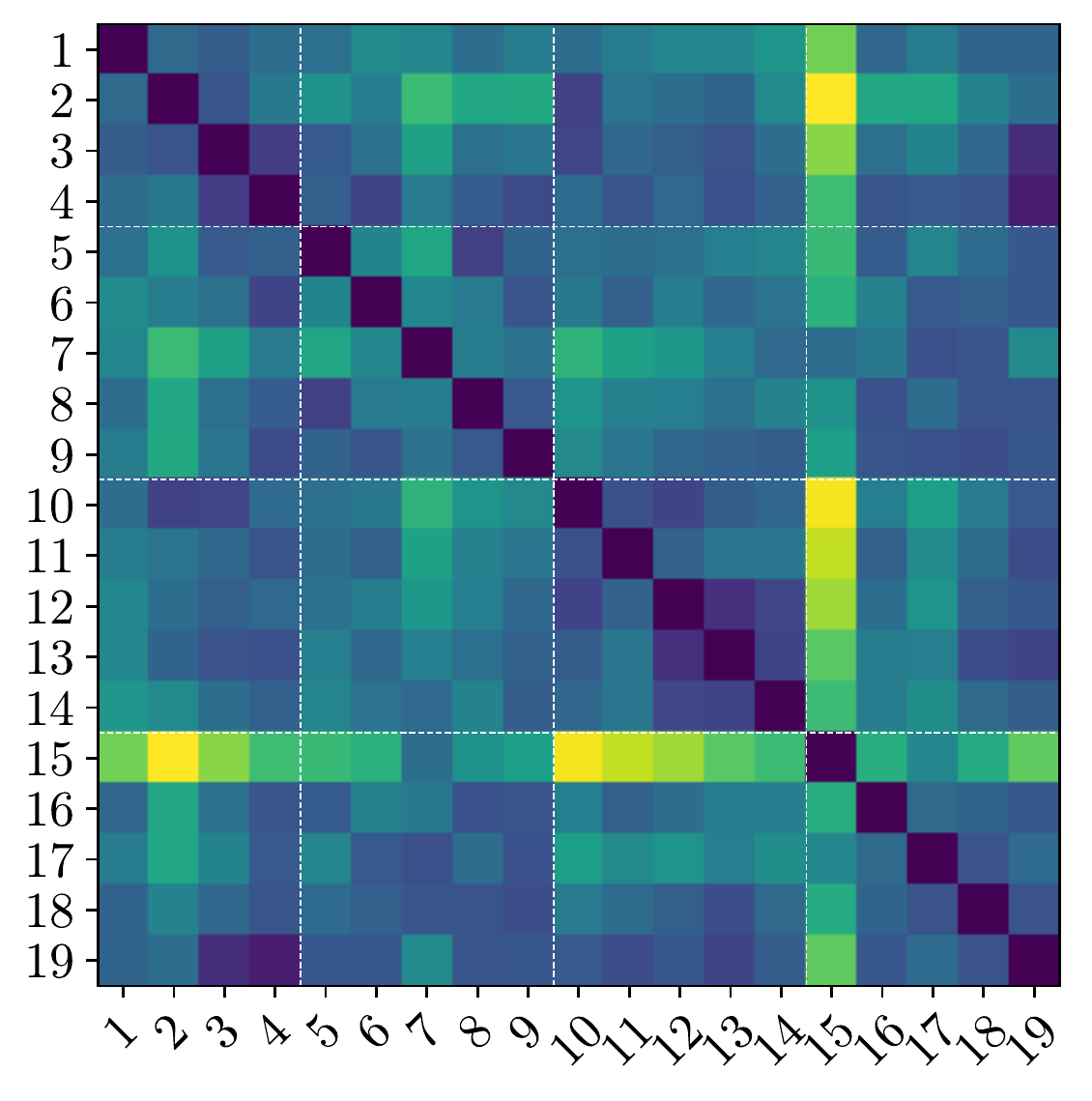}}
\\
\multicolumn{3}{c}{\includegraphics[width = 5.5cm]{img/general/temp_bar.pdf}} \\
\end{tabular}
\caption{MNIST neural networks' Persistent Homology distance matrices means.}
\label{fig:ap_mnist_means}
\end{figure}

\begin{figure}[H]
\centering
\begin{tabular}{ccc}
\subfloat[Heat distance.]{\includegraphics[width = 2in]{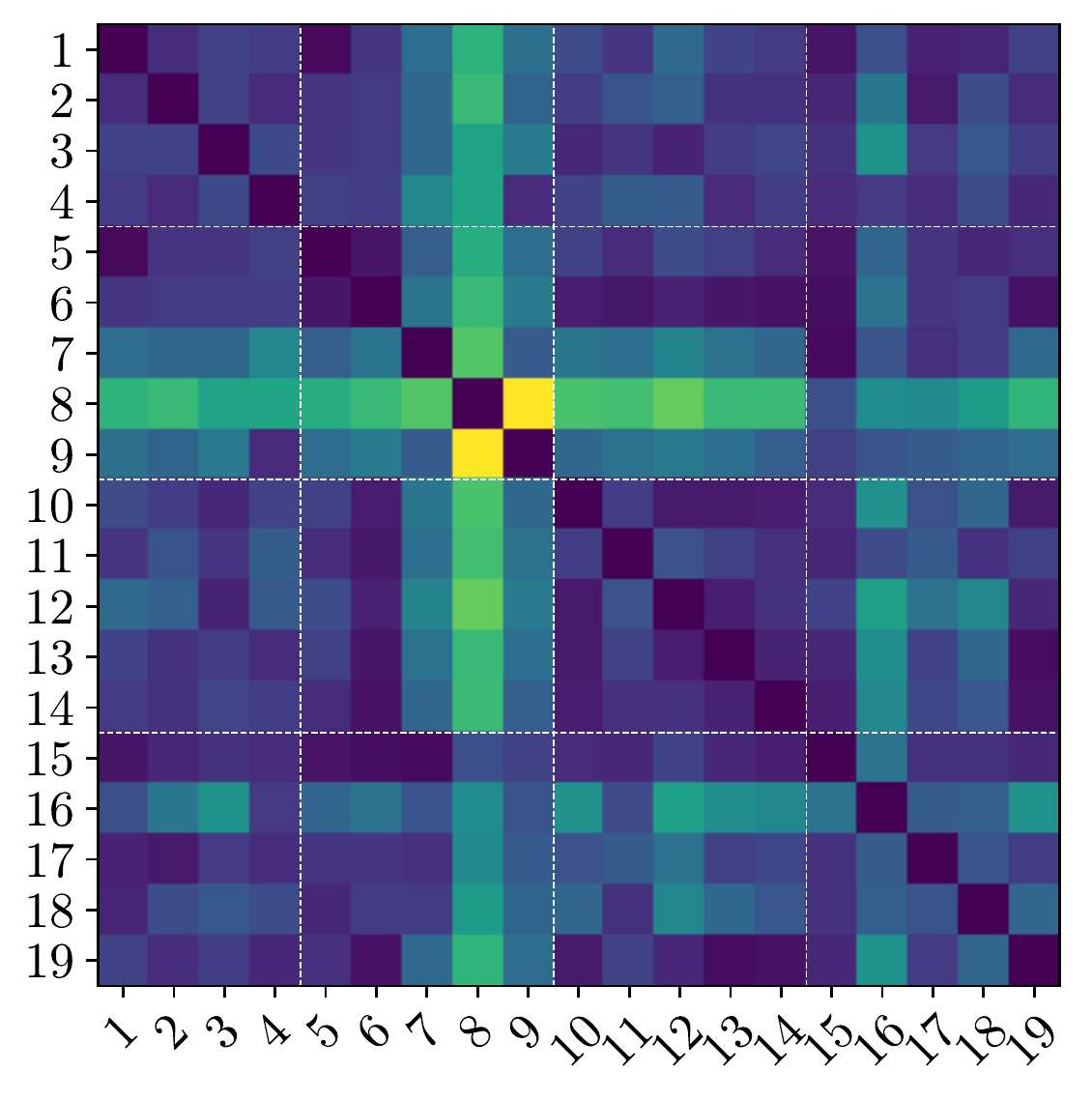}} &
\subfloat[Silhouette distance.]{\includegraphics[width = 2in]{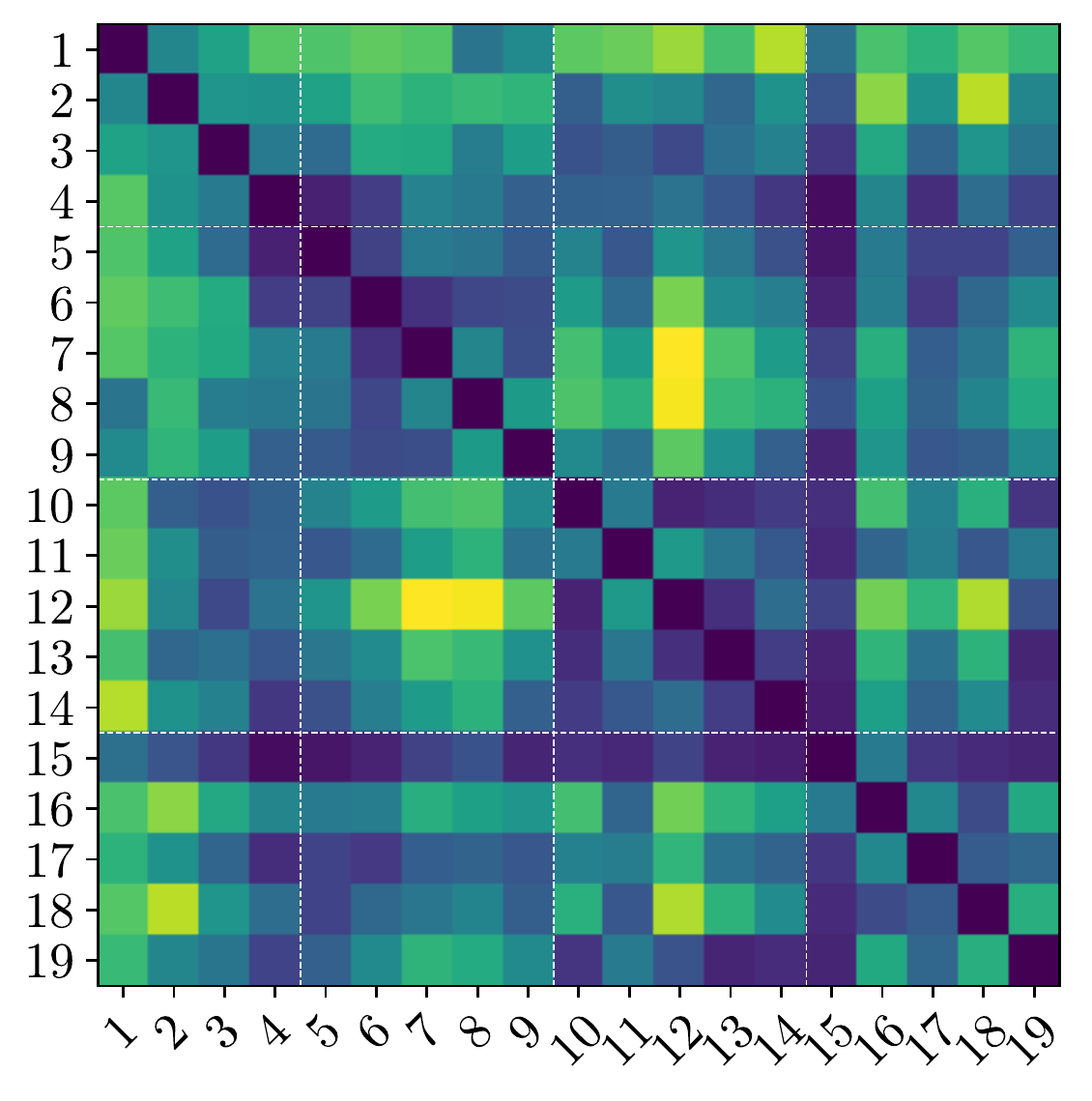}} &
\subfloat[Landscape distance.]{\includegraphics[width = 2in]{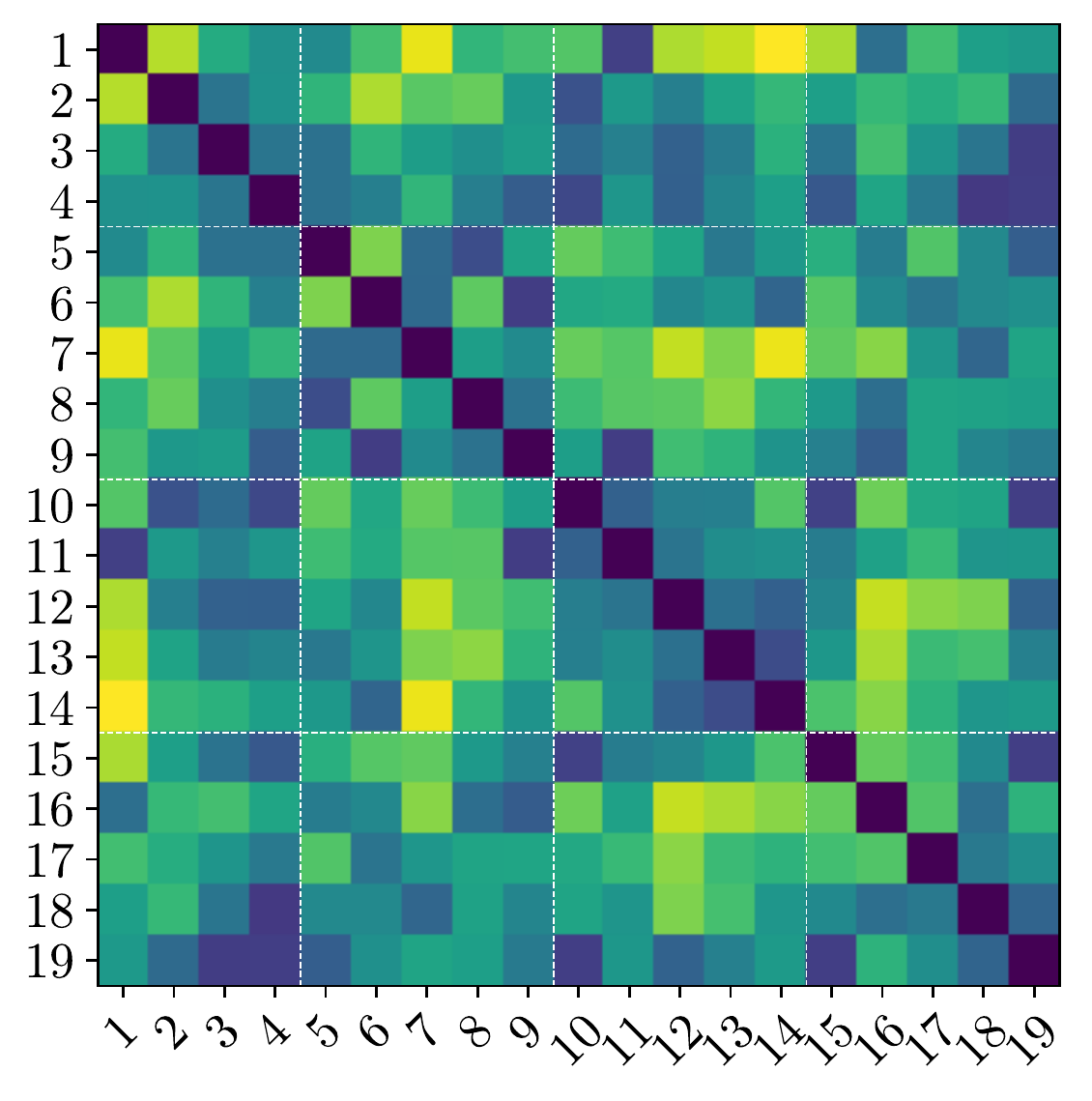}}
\\
\multicolumn{3}{c}{\includegraphics[width = 5.5cm]{img/general/temp_bar.pdf}} \\
\end{tabular}
\caption{MNIST neural networks' Persistent Homology distance matrices standard deviations.}
\label{fig:ap_mnist_std}
\end{figure}

\begin{figure}[H]
\centering
\begin{tabular}{ccc}
\subfloat[Heat distance.]{\includegraphics[width = 2in]{img/results/fashion_mnist/heat_distance_matrices.pdf}} &
\subfloat[Silhouette distance.]{\includegraphics[width = 2in]{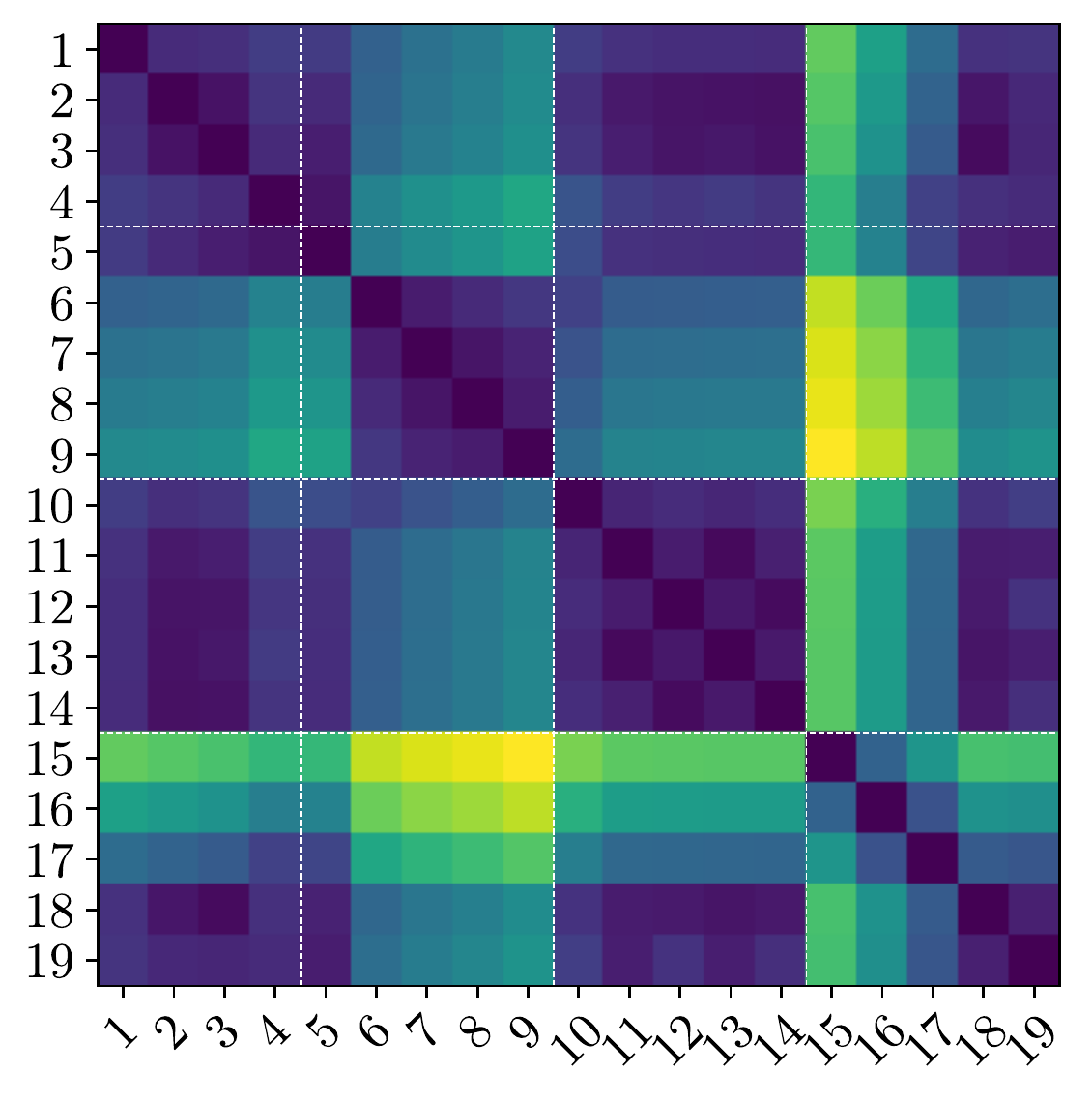}} &
\subfloat[Landscape distance.]{\includegraphics[width = 2in]{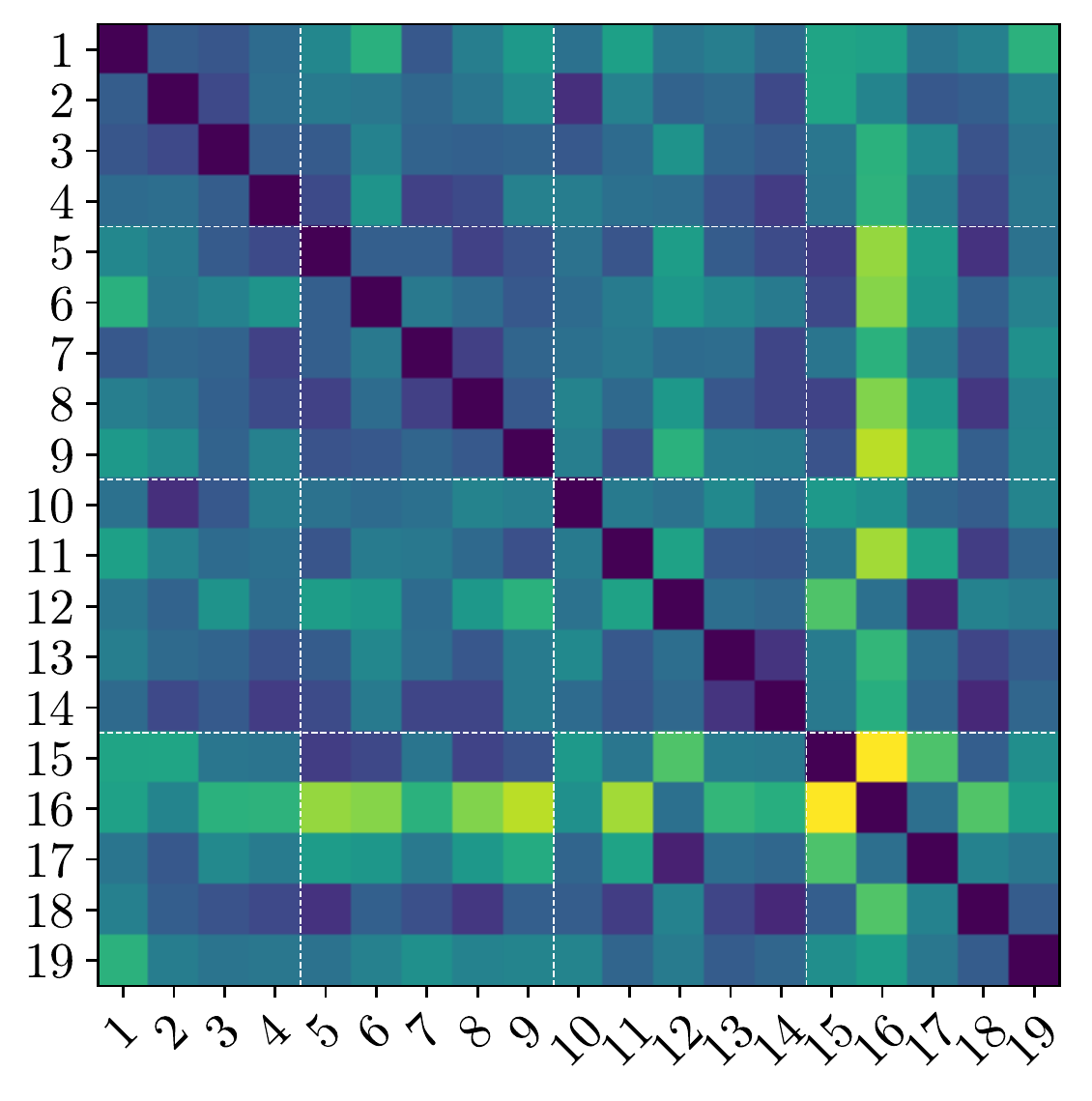}}
\\
\multicolumn{3}{c}{\includegraphics[width = 5.5cm]{img/general/temp_bar.pdf}} \\
\end{tabular}
\caption{Fashion MNIST neural networks' Persistent Homology distance matrices means.}
\label{fig:ap_fashion_mnist_mean}
\end{figure}

\begin{figure}[H]
\centering
\begin{tabular}{ccc}
\subfloat[Heat distance.]{\includegraphics[width = 2in]{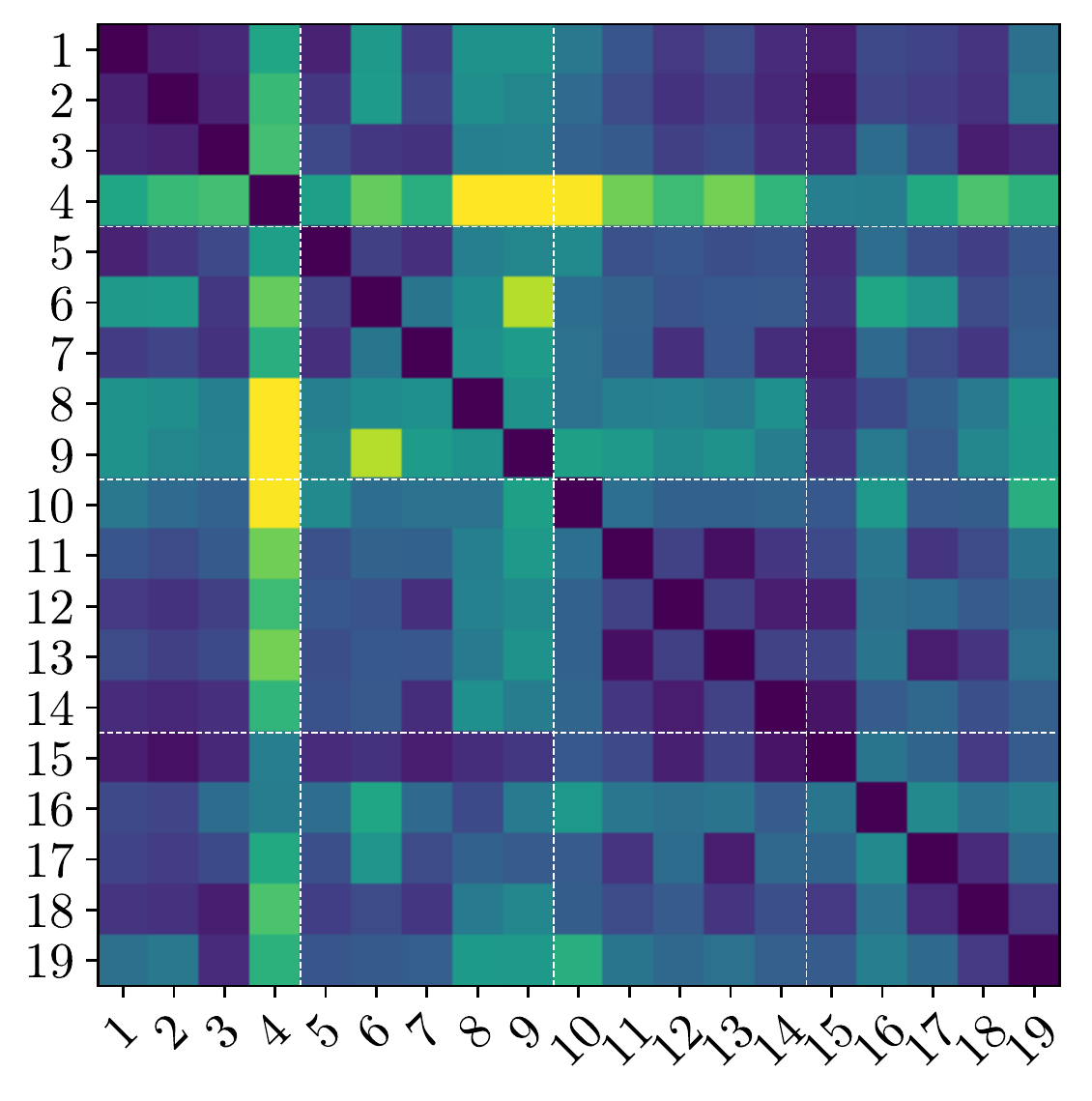}} &
\subfloat[Silhouette distance.]{\includegraphics[width = 2in]{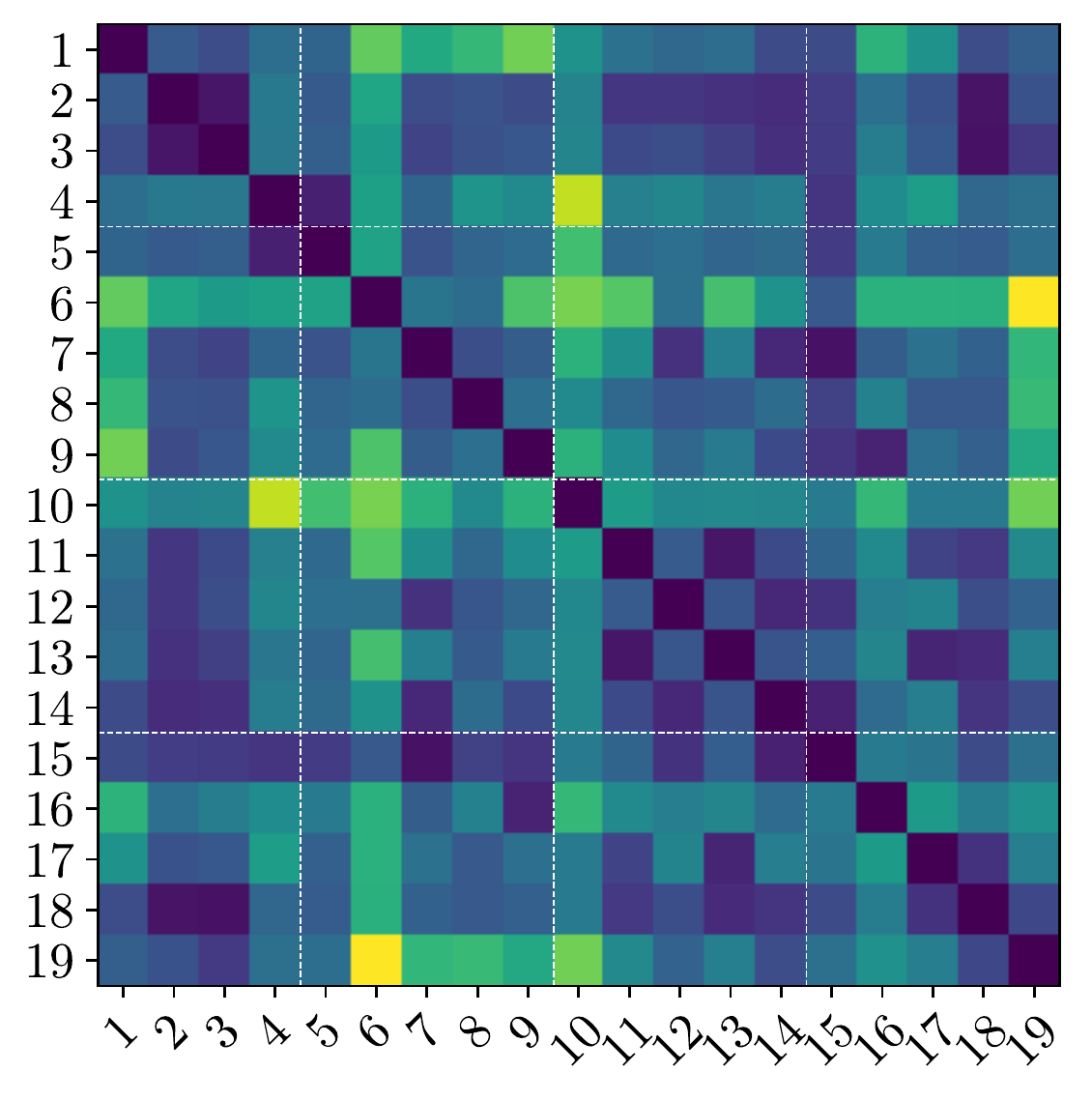}} &
\subfloat[Landscape distance.]{\includegraphics[width = 2in]{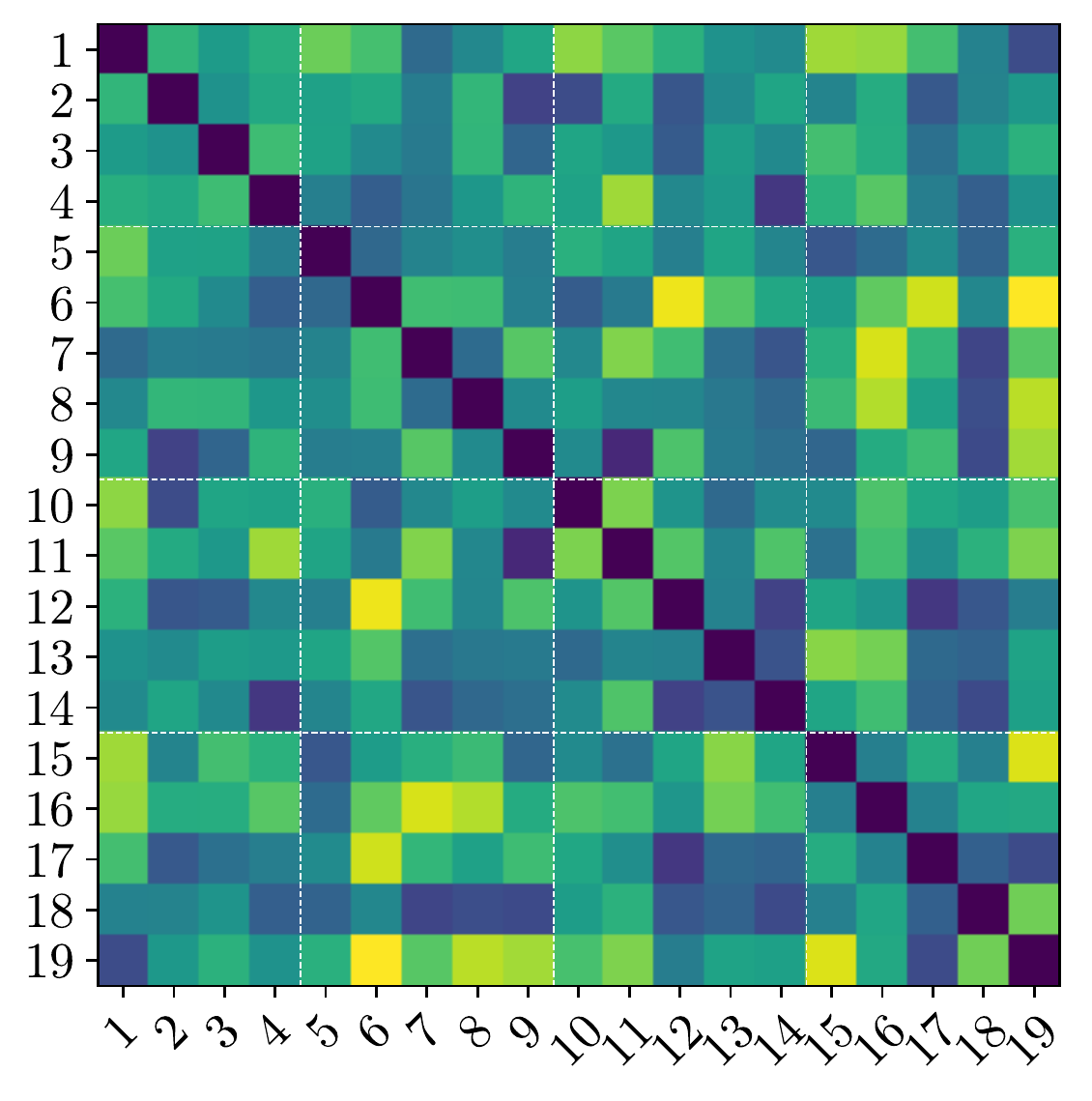}}
\\
\multicolumn{3}{c}{\includegraphics[width = 5.5cm]{img/general/temp_bar.pdf}} \\
\end{tabular}
\caption{Fashion MNIST neural networks' Persistent Homology distance matrices standard deviations.}
\label{fig:ap_fashion_mnist_std}
\end{figure}

\begin{figure}[H]
\centering
\begin{tabular}{ccc}
\subfloat[Heat distance.]{\includegraphics[width = 2in]{img/results/cifar_10/heat_distance_matrices.pdf}} &
\subfloat[Silhouette distance.]{\includegraphics[width = 2in]{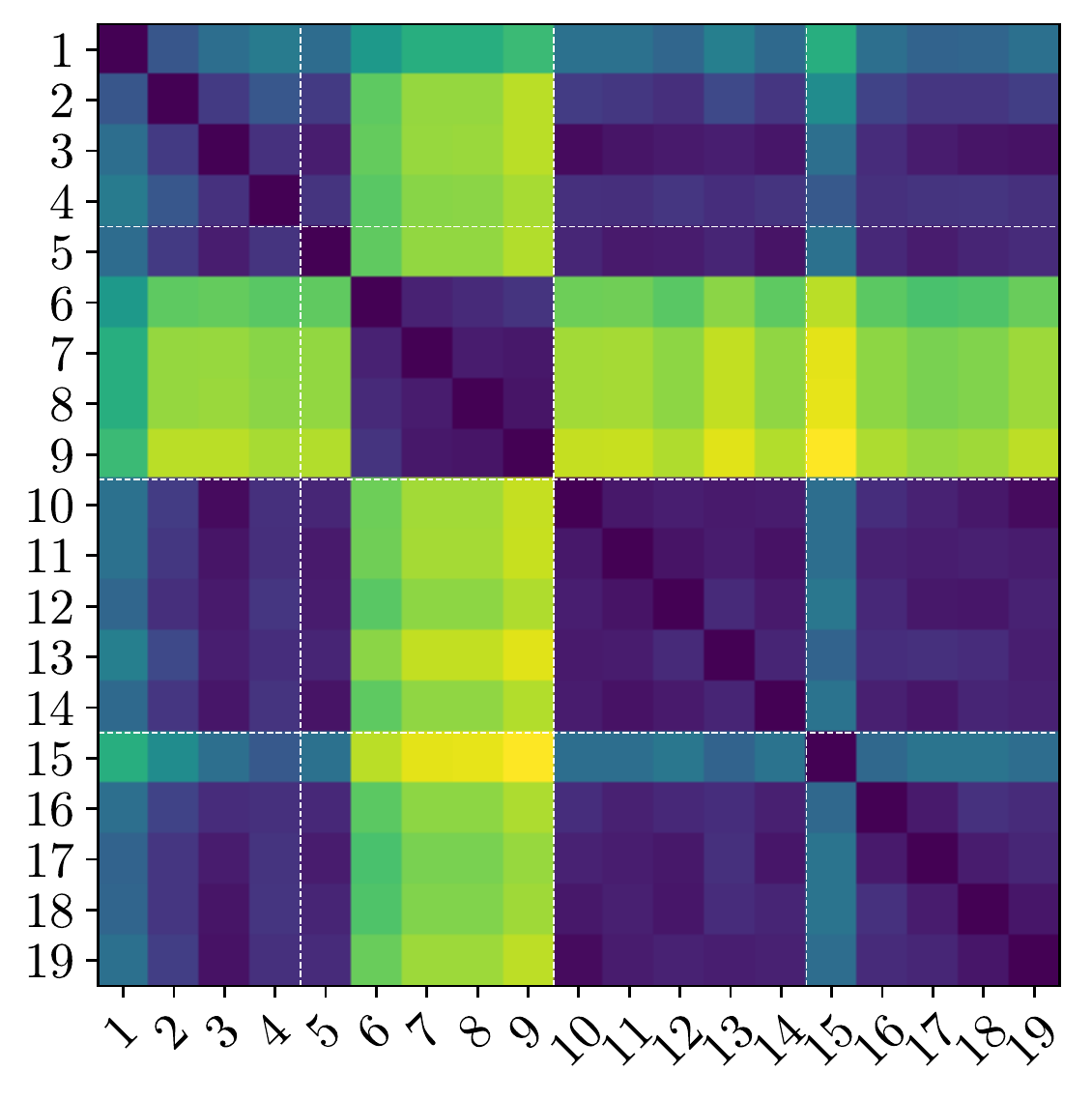}} &
\subfloat[Landscape distance.]{\includegraphics[width = 2in]{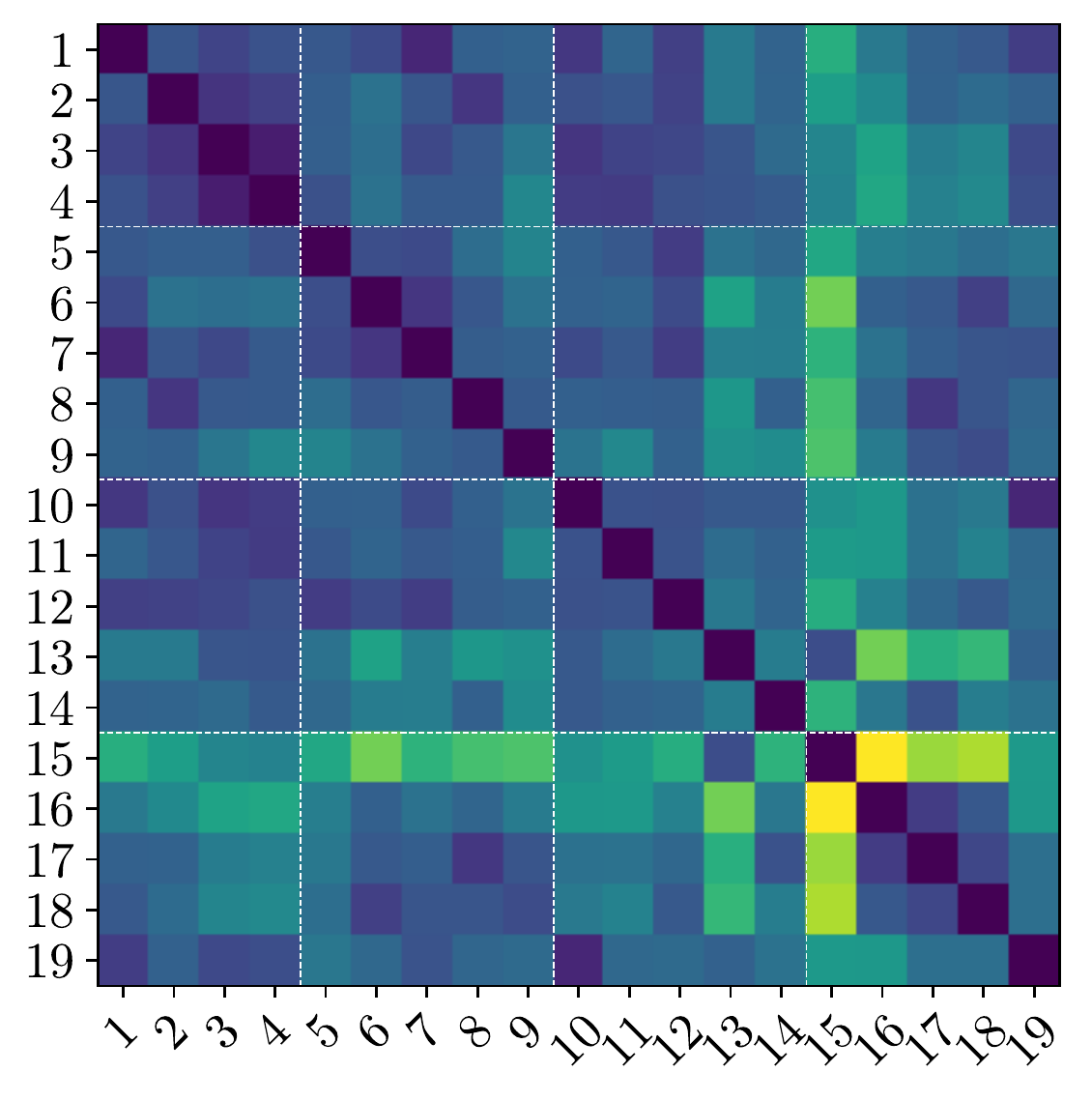}}
\\
\multicolumn{3}{c}{\includegraphics[width = 5.5cm]{img/general/temp_bar.pdf}} \\
\end{tabular}
\caption{CIFAR-10 neural networks' Persistent Homology distance matrices means.}
\label{fig:ap_cifar_10_mean}
\end{figure}

\begin{figure}[H]
\centering
\begin{tabular}{ccc}
\subfloat[Heat distance.]{\includegraphics[width = 2in]{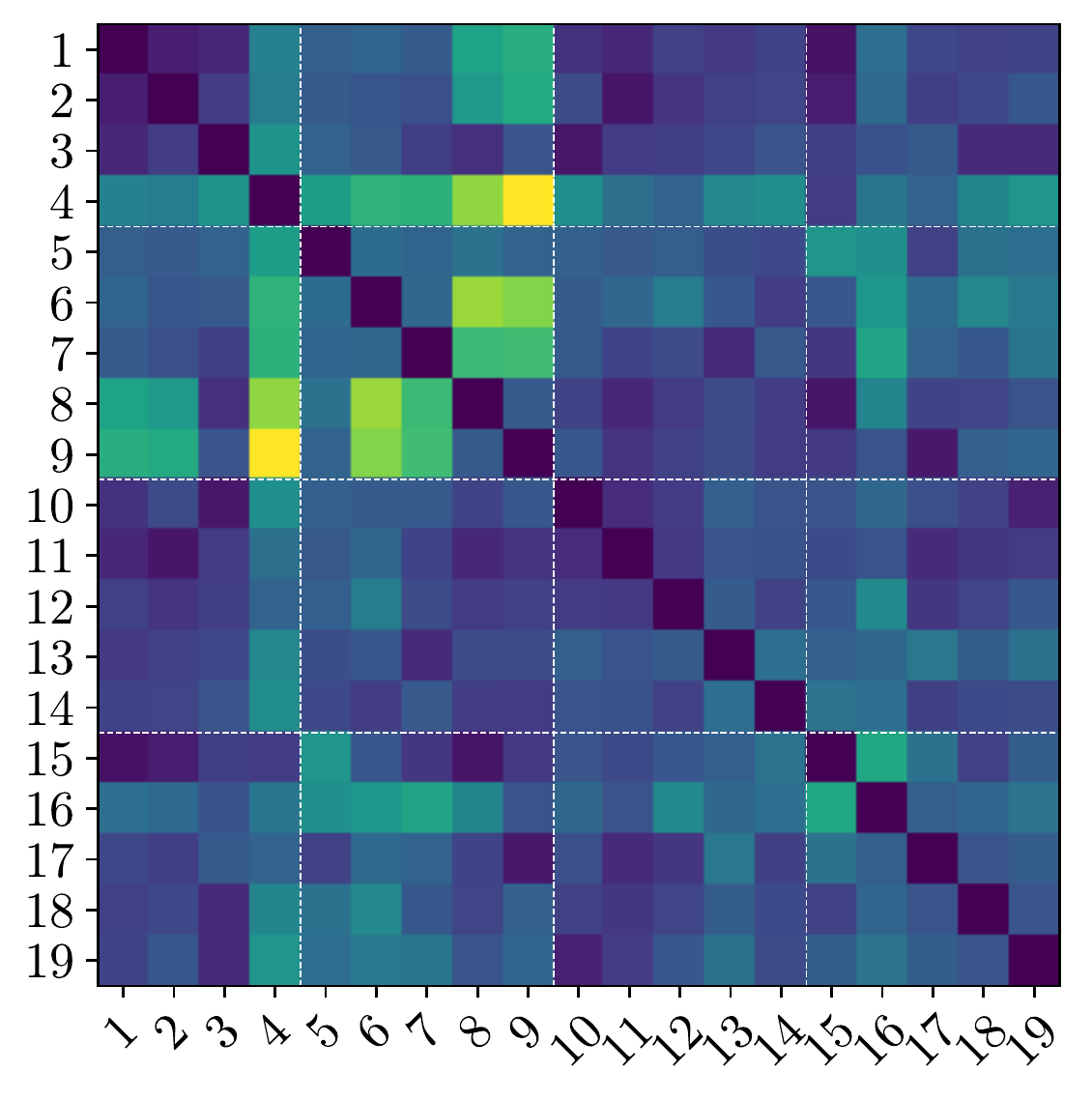}} &
\subfloat[Silhouette distance.]{\includegraphics[width = 2in]{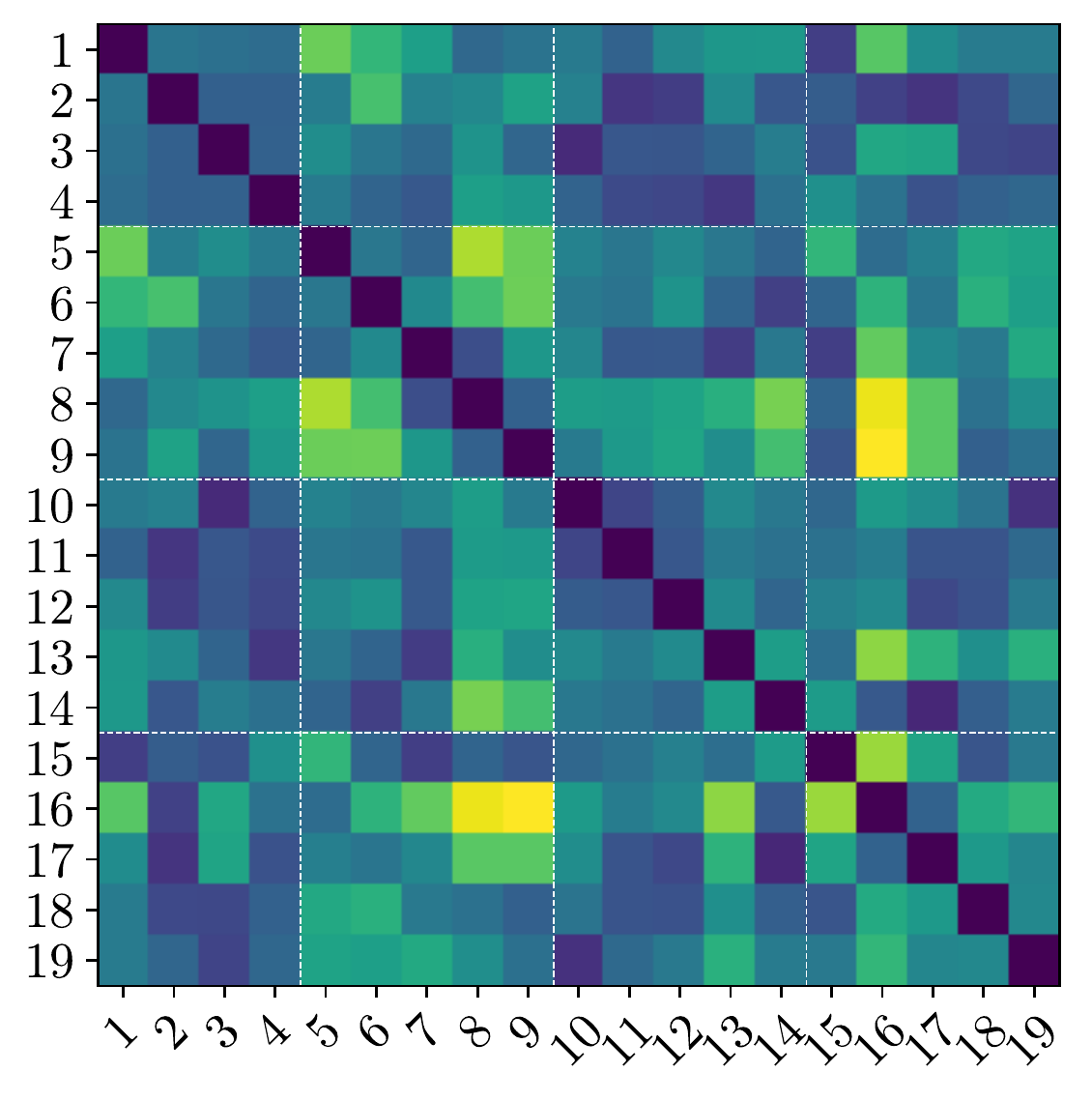}} &
\subfloat[Landscape distance.]{\includegraphics[width = 2in]{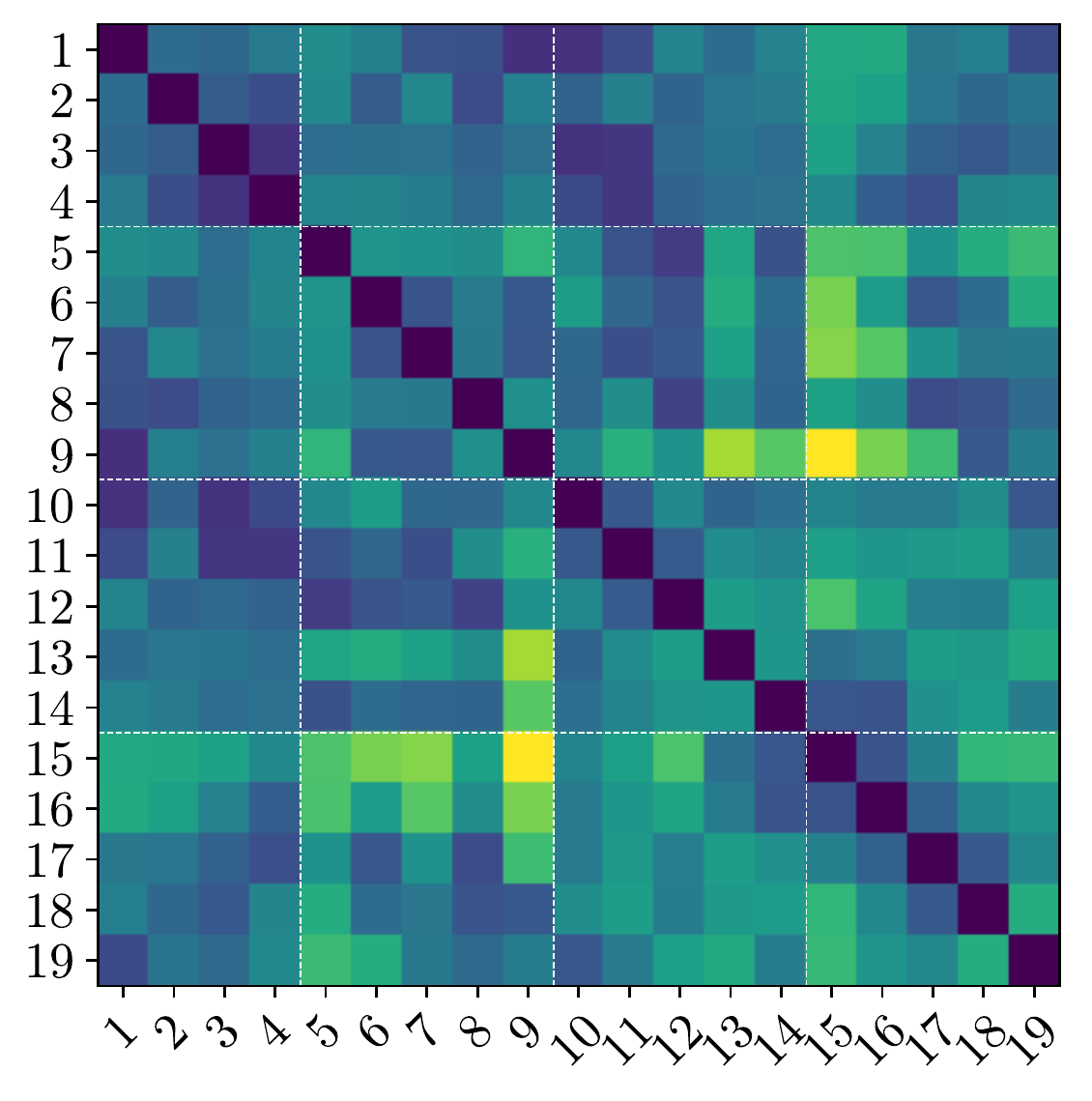}}
\\
\multicolumn{3}{c}{\includegraphics[width = 5.5cm]{img/general/temp_bar.pdf}} \\
\end{tabular}
\caption{CIFAR-10 neural networks' Persistent Homology distance matrices standard deviations.}
\label{fig:ap_cifar_10_std}
\end{figure}

\begin{figure}[H]
\centering
\begin{tabular}{ccc}
\subfloat[Heat distance.]{\includegraphics[width = 2in]{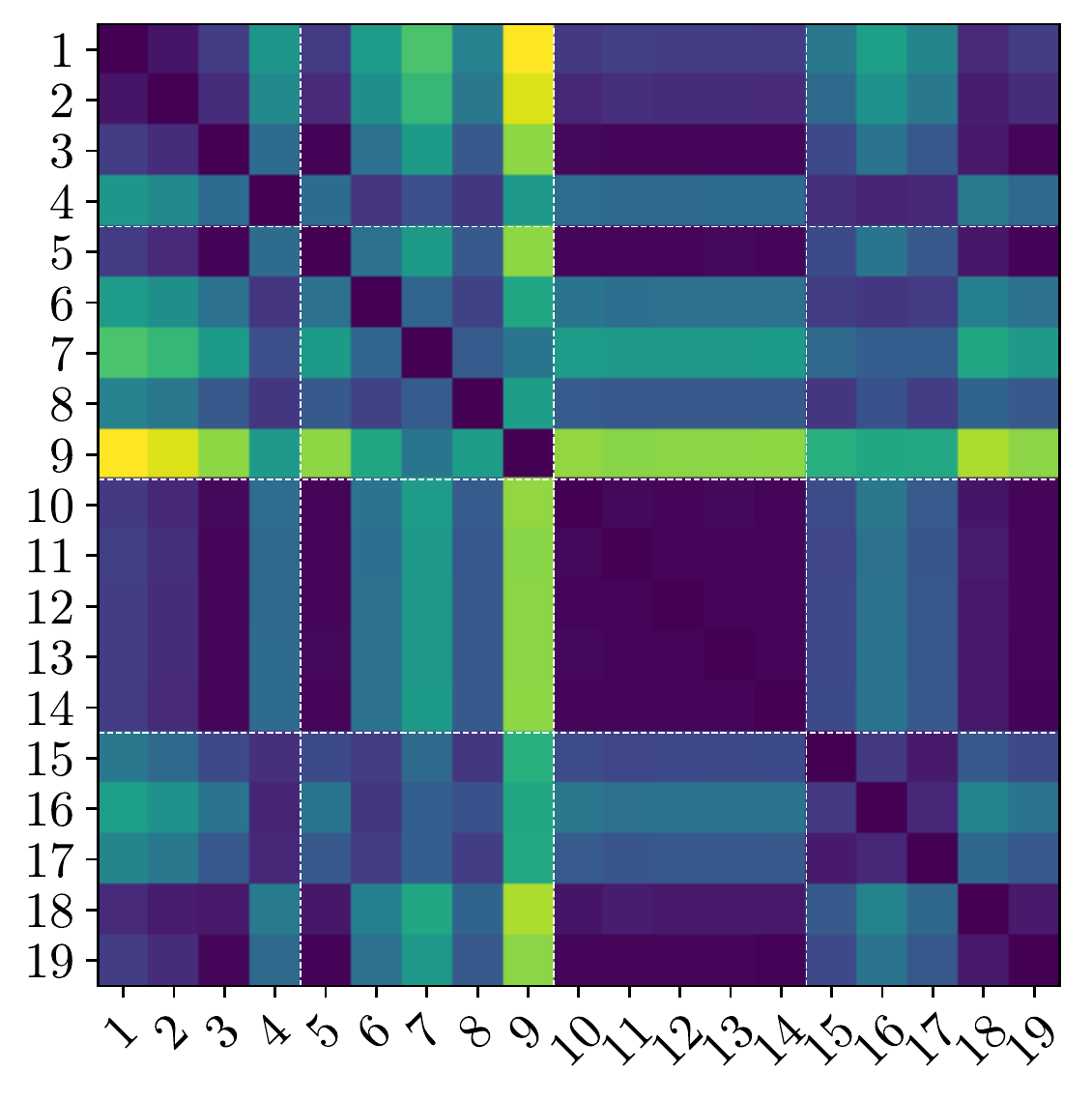}} &
\subfloat[Silhouette distance.]{\includegraphics[width = 2in]{img/results/language_identification/silhouette_distance_matrices.pdf}} &
\subfloat[Landscape distance.]{\includegraphics[width = 2in]{img/results/language_identification/landscape_distance_matrices.pdf}}
\\
\multicolumn{3}{c}{\includegraphics[width = 5.5cm]{img/general/temp_bar.pdf}} \\
\end{tabular}
\caption{Language Identification neural networks' Persistent Homology distance matrices means.}
\label{fig:ap_language_identification_mean}
\end{figure}

\begin{figure}[H]
\centering
\begin{tabular}{ccc}
\subfloat[Heat distance.]{\includegraphics[width = 2in]{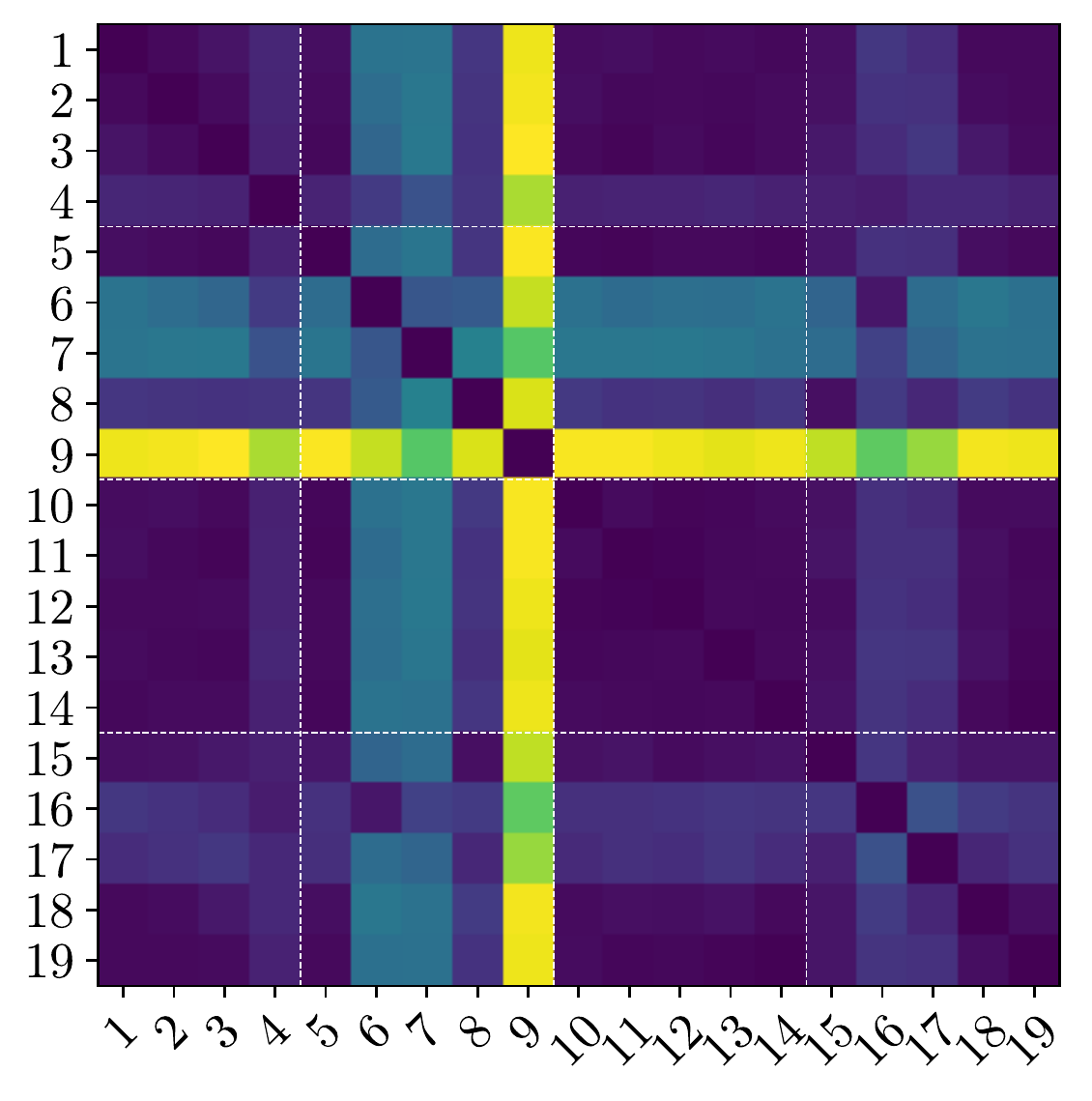}} &
\subfloat[Silhouette distance.]{\includegraphics[width = 2in]{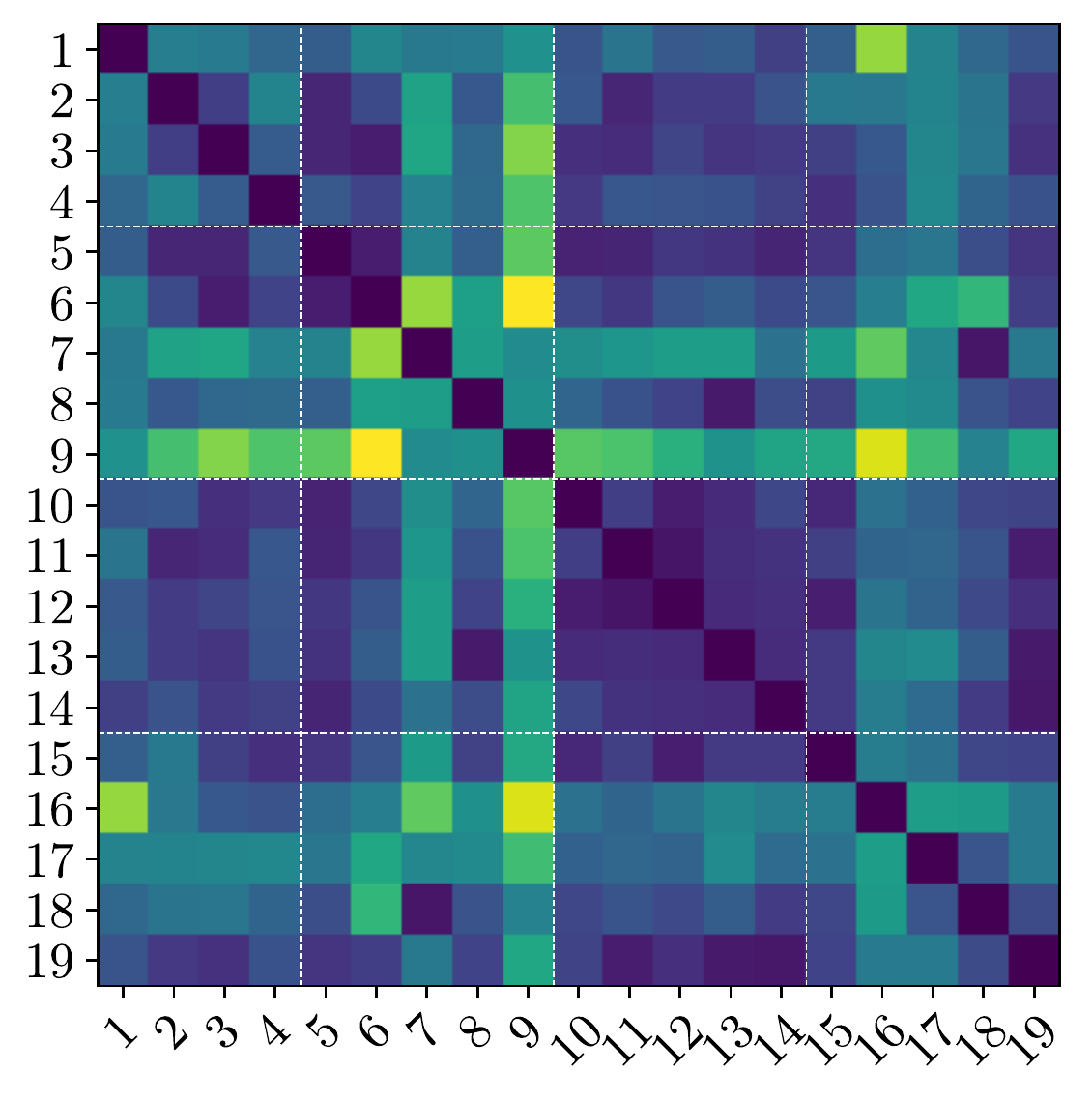}} &
\subfloat[Landscape distance.]{\includegraphics[width = 2in]{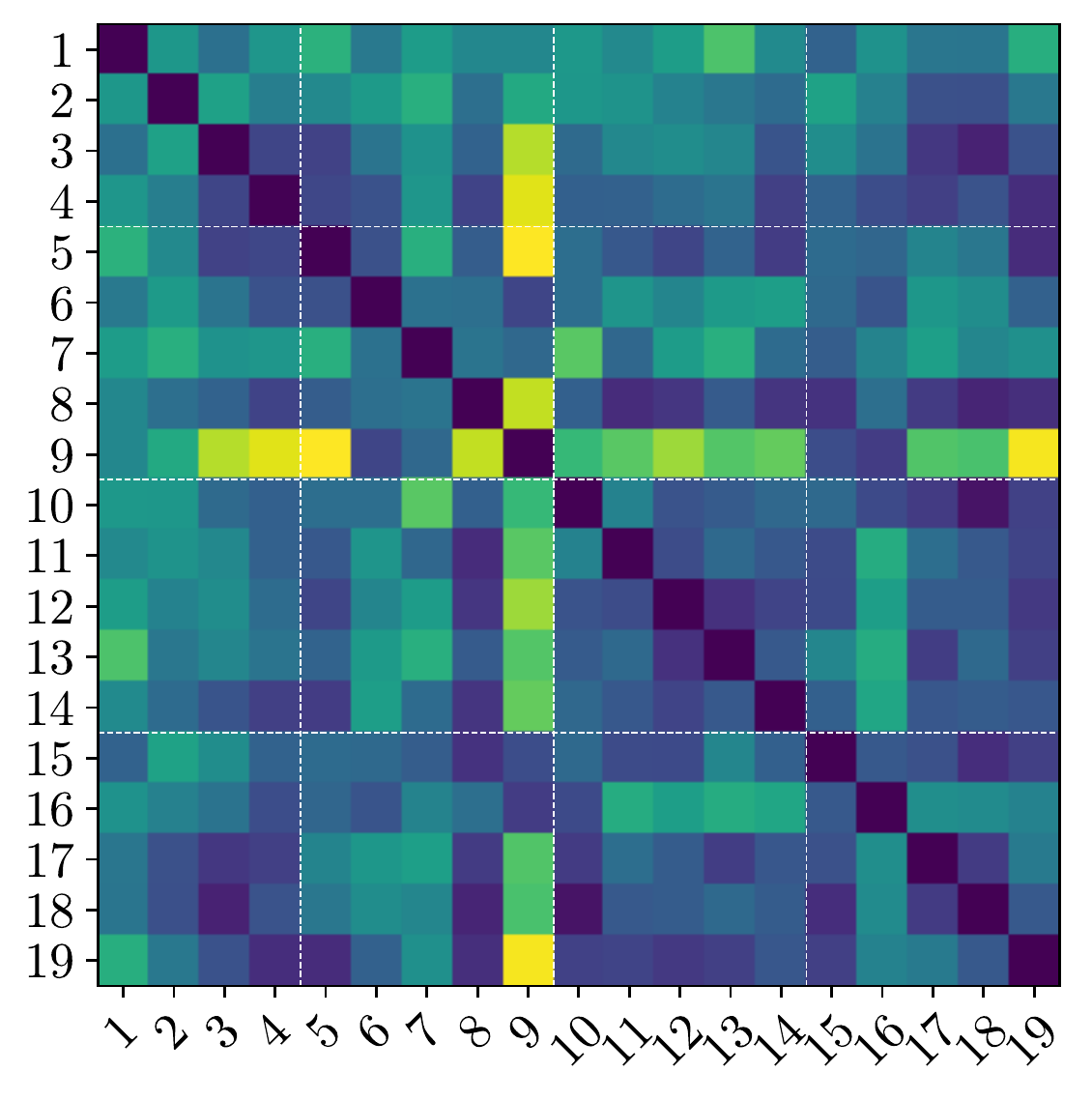}}
\\
\multicolumn{3}{c}{\includegraphics[width = 5.5cm]{img/general/temp_bar.pdf}} \\
\end{tabular}
\caption{Language Identification neural networks' Persistent Homology distance matrices standard deviations.}
\label{fig:ap_language_identification_std}
\end{figure}

\begin{figure}[H]
\centering
\begin{tabular}{ccc}
\subfloat[Heat distance.]{\includegraphics[width = 2in]{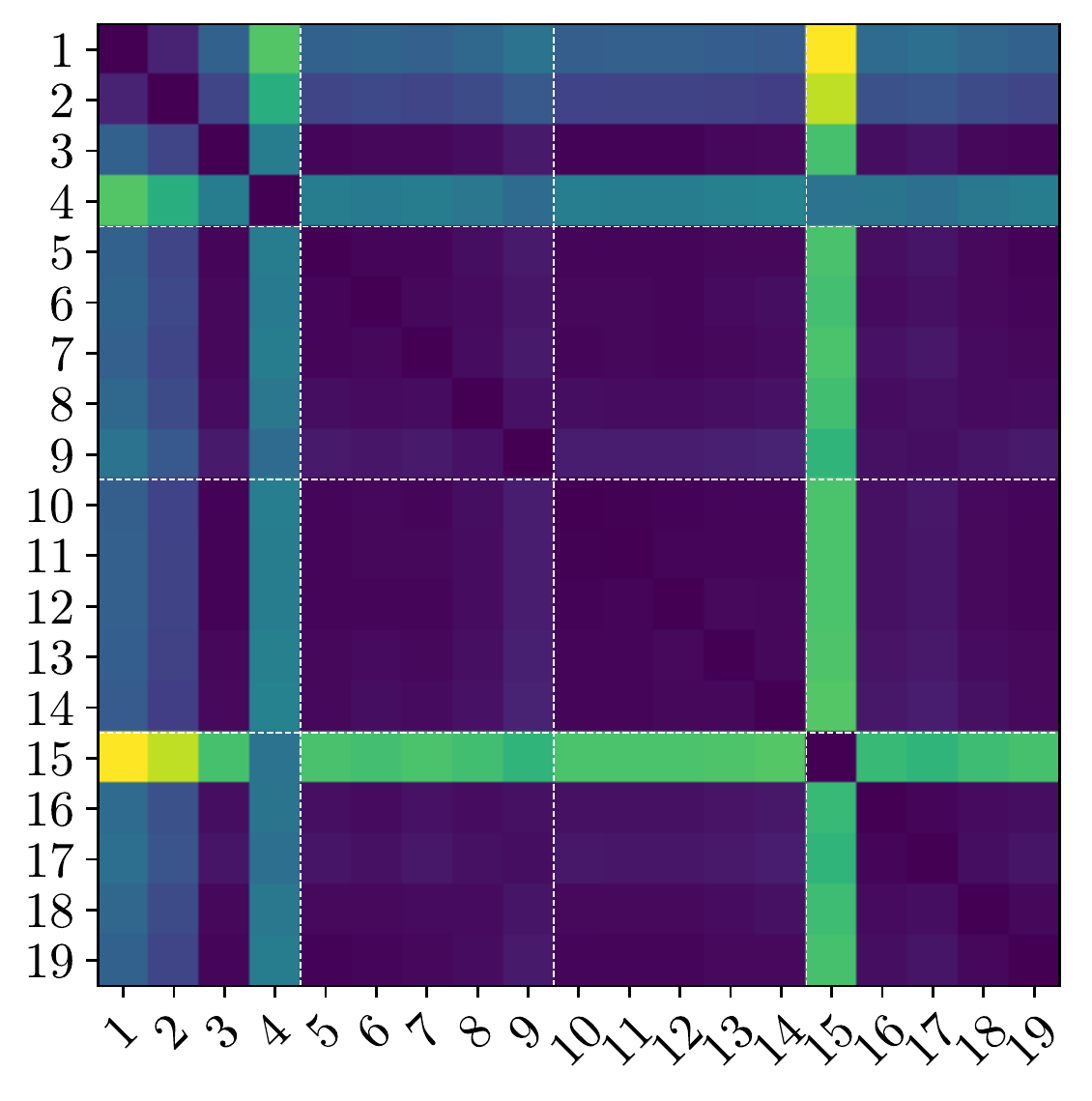}} &
\subfloat[Silhouette distance.]{\includegraphics[width = 2in]{img/results/reuters/silhouette_distance_matrices.pdf}} &
\subfloat[Landscape distance.]{\includegraphics[width = 2in]{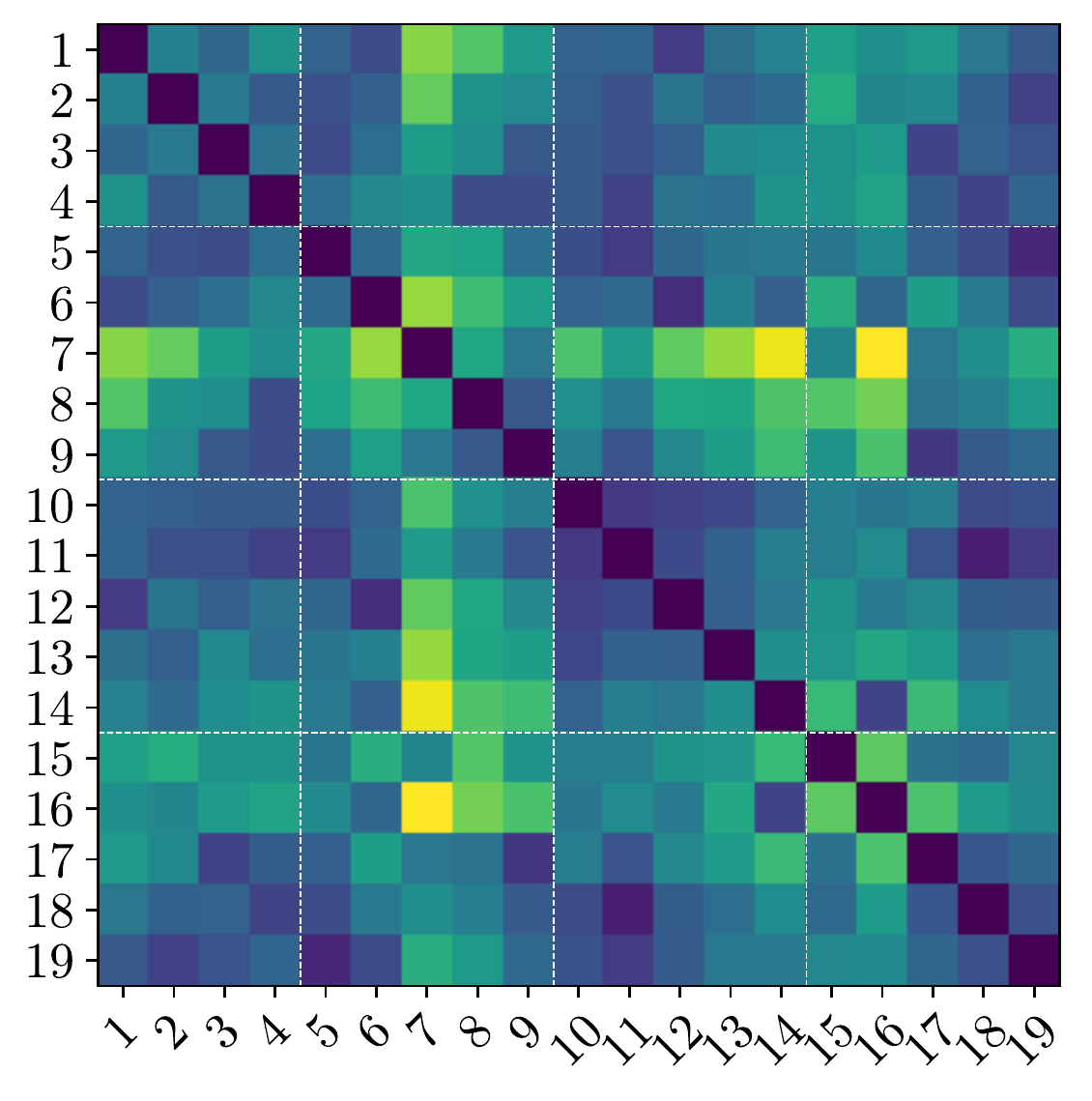}}
\\
\multicolumn{3}{c}{\includegraphics[width = 5.5cm]{img/general/temp_bar.pdf}} \\
\end{tabular}
\caption{Reuters neural networks' Persistent Homology distance matrices means.}
\label{fig:ap_reuters_mean}
\end{figure}

\begin{figure}[H]
\centering
\begin{tabular}{ccc}
\subfloat[Heat distance.]{\includegraphics[width = 2in]{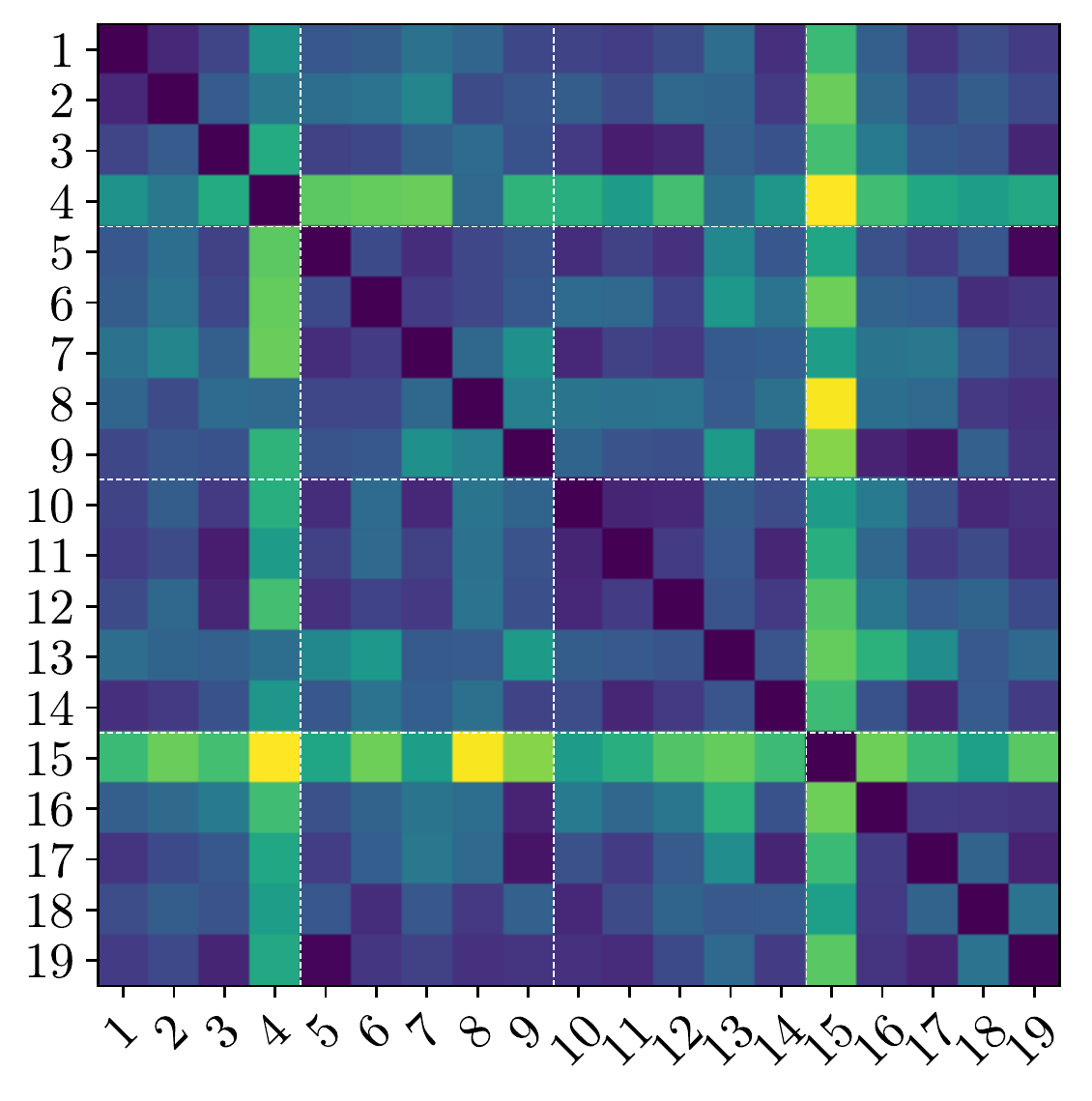}} &
\subfloat[Silhouette distance.]{\includegraphics[width = 2in]{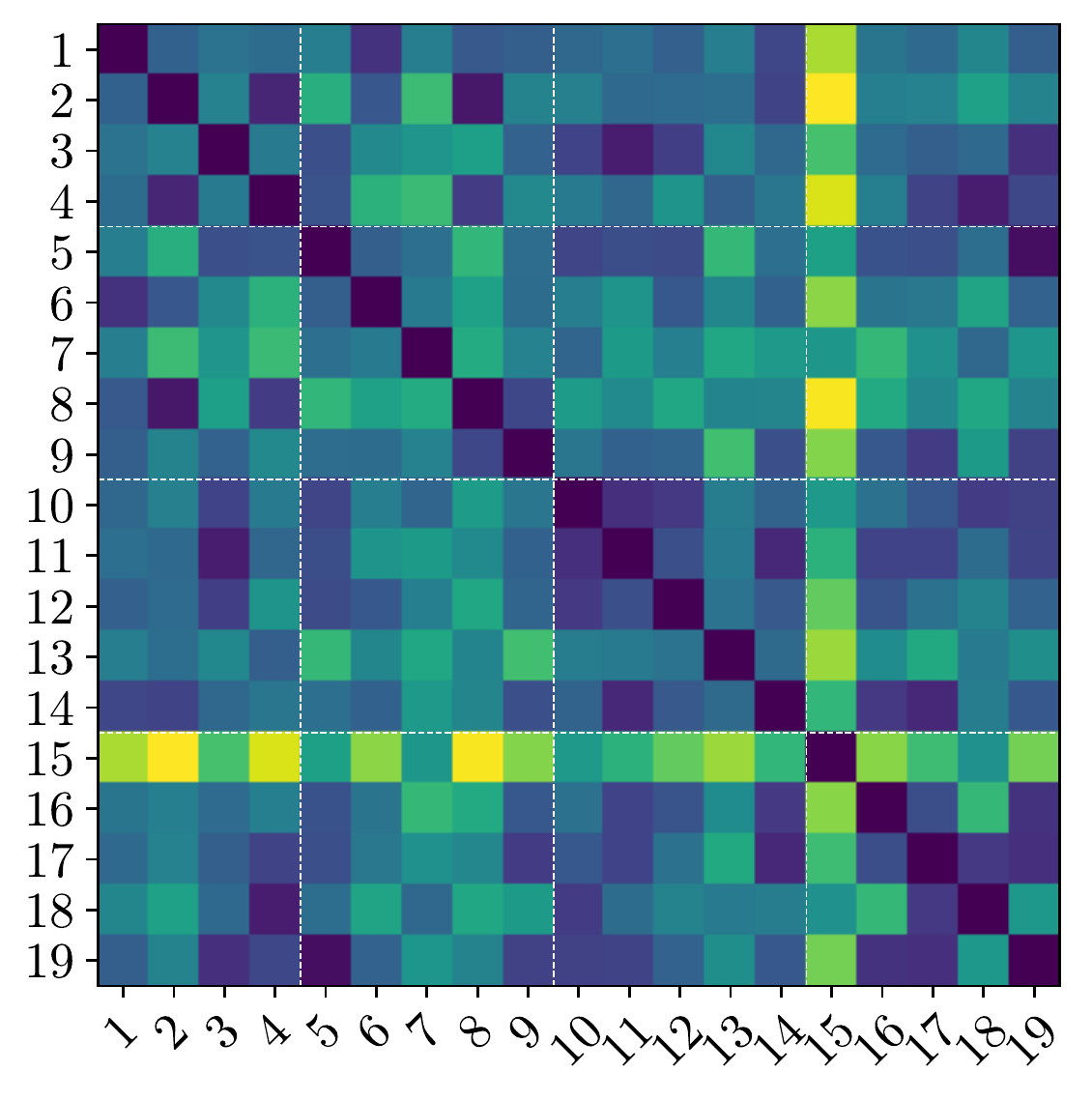}} &
\subfloat[Landscape distance.]{\includegraphics[width = 2in]{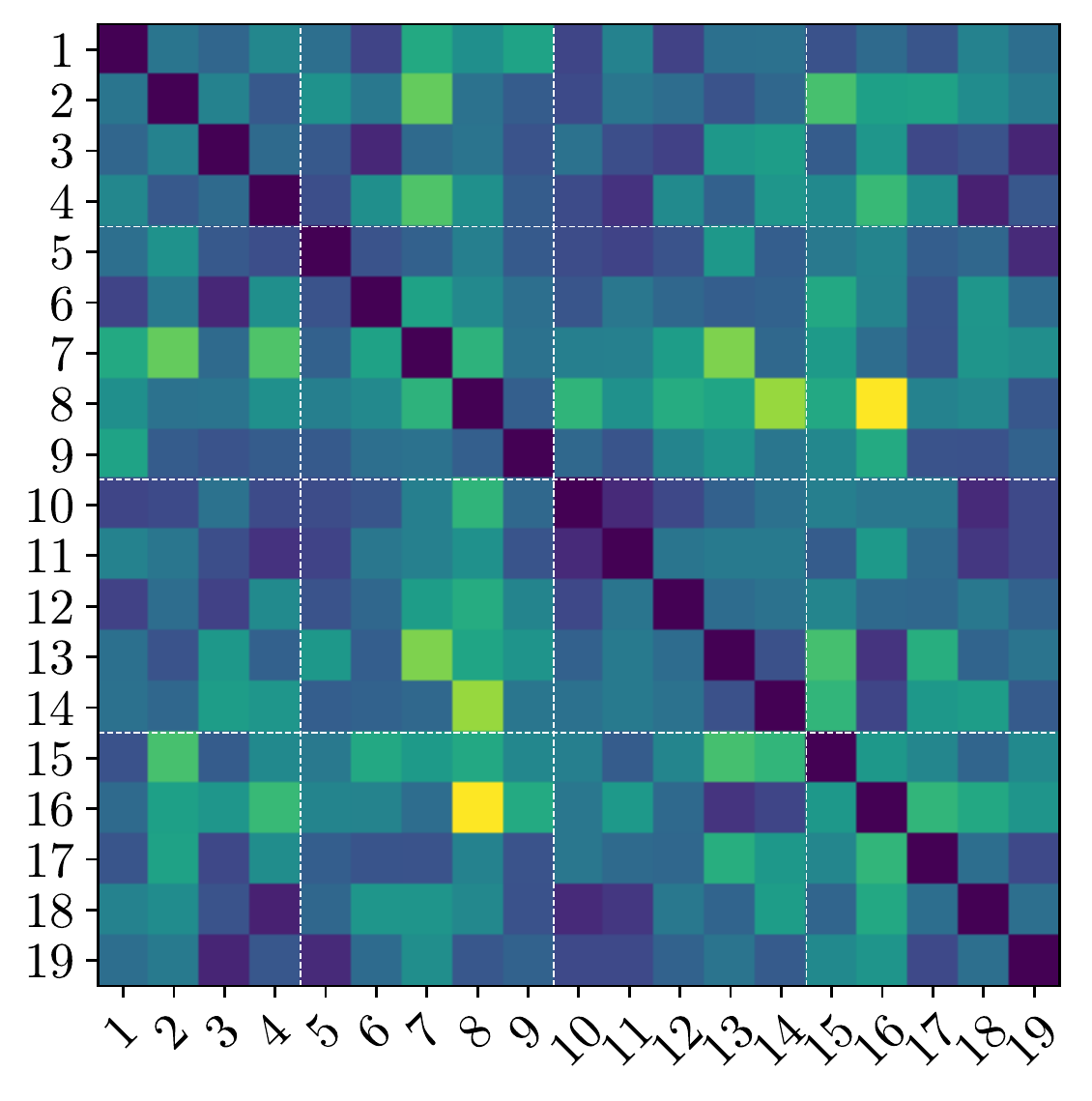}}
\\
\multicolumn{3}{c}{\includegraphics[width = 5.5cm]{img/general/temp_bar.pdf}} \\
\end{tabular}
\caption{Reuters neural networks' Persistent Homology distance matrices standard deviations.}
\label{fig:ap_reuters_std}
\end{figure}

\begin{table}[H]
\centering
\begin{tabular}{@{}llrr@{}}
\toprule
Discretization & Experiment & Mean & Standard deviation \\ \midrule
Heat & Layer size & 0.5128 & 0.3391 \\
Heat & Number layers & 0.3633 & 0.1933 \\
Heat & Input order & 0.0291 & 0.0100 \\
Heat & Number labels & 0.2279 & 0.1352 \\
Landscape & Layer size & 0.3077 & 0.0727 \\
Landscape & Number layers & 0.3776 & 0.1096 \\
Landscape & Input order & 0.2719 & 0.0822 \\
Landscape & Number labels & 0.4266 & 0.1705 \\
Silhouette & Layer size & 0.2110 & 0.0552 \\
Silhouette & Number layers & 0.2681 & 0.1953 \\
Silhouette & Input order & 0.1115 & 0.0364 \\
Silhouette & Number labels & 0.4126 & 0.2270 \\
 \bottomrule
\end{tabular}
\caption{MNIST statistics of experiment groups.}
\label{tab:mnist-statistics}
\end{table}

\begin{table}[H]
\centering
\begin{tabular}{@{}llrr@{}}
\toprule
Discretization & Experiment & Mean & Standard deviation \\ \midrule
Heat & Layer size & 0.5380 & 0.3487 \\
Heat & Number layers & 0.1774 & 0.0739 \\
Heat & Input order & 0.0308 & 0.0132 \\
Heat & Number labels & 0.2679 & 0.1427 \\
Landscape & Layer size & 0.2976 & 0.0482 \\
Landscape & Number layers & 0.2886 & 0.0631 \\
Landscape & Input order & 0.3583 & 0.1114 \\
Landscape & Number labels & 0.5301 & 0.2149 \\
Silhouette & Layer size & 0.1265 & 0.0409 \\
Silhouette & Number layers & 0.2592 & 0.2024 \\
Silhouette & Input order & 0.0824 & 0.0353 \\
Silhouette & Number labels & 0.4150 & 0.1940 \\
 \bottomrule
\end{tabular}
\caption{Fashion MNIST statistics of experiment groups.}
\label{tab:fashion-mnist-statistics}
\end{table}

\begin{table}[H]
\centering
\begin{tabular}{@{}llrr@{}}
\toprule
Discretization & Experiment & \multicolumn{1}{l}{Mean} & \multicolumn{1}{l}{Standard deviation} \\ \midrule
Heat & Layer size & 0.5414 & 0.3319 \\
Heat & Number layers & 0.1314 & 0.0529 \\
Heat & Input order & 0.0243 & 0.0068 \\
Heat & Number labels & 0.1084 & 0.0870 \\
Landscape & Layer size & 0.1907 & 0.0630 \\
Landscape & Number layers & 0.2967 & 0.0789 \\
Landscape & Input order & 0.3119 & 0.0584 \\
Landscape & Number labels & 0.5193 & 0.2803 \\
Silhouette & Layer size & 0.2705 & 0.0964 \\
Silhouette & Number layers & 0.3885 & 0.3626 \\
Silhouette & Input order & 0.0769 & 0.0204 \\
Silhouette & Number labels & 0.2049 & 0.1341 \\
 \bottomrule
\end{tabular}
\caption{CIFAR-10 statistics of experiment groups.}
\label{tab:cifar-statistics}
\end{table}

\begin{table}[H]
\centering
\begin{tabular}{@{}llrr@{}}
\toprule
Discretization & Experiment & Mean & Standard deviation \\ \midrule
Heat & Layer size & 0.2856 & 0.1771 \\
Heat & Number layers & 0.4396 & 0.1811 \\
Heat & Input order & 0.0159 & 0.0040 \\
Heat & Number labels & 0.2374 & 0.1243 \\
Landscape & Layer size & 0.3246 & 0.1254 \\
Landscape & Number layers & 0.5643 & 0.1643 \\
Landscape & Input order & 0.2320 & 0.0982 \\
Landscape & Number labels & 0.3519 & 0.1671 \\
Silhouette & Layer size & 0.1501 & 0.0384 \\
Silhouette & Number layers & 0.3374 & 0.1420 \\
Silhouette & Input order & 0.0699 & 0.0159 \\
Silhouette & Number labels & 0.3737 & 0.1719 \\
 \bottomrule
\end{tabular}
\caption{Language Identification statistics of experiment groups.}
\label{tab:language-statistics}
\end{table}

\begin{table}[H]
\centering
\begin{tabular}{@{}llrr@{}}
\toprule
Discretization & Experiment & Mean & Standard deviation \\ \midrule
Heat & Layer size & 0.4004 & 0.2220 \\
Heat & Number layers & 0.0412 & 0.0211 \\
Heat & Input order & 0.0166 & 0.0051 \\
Heat & Number labels & 0.2927 & 0.3185 \\
Landscape & Layer size & 0.3950 & 0.0733 \\
Landscape & Number layers & 0.5265 & 0.1678 \\
Landscape & Input order & 0.3060 & 0.1028 \\
Landscape & Number labels & 0.4534 & 0.1651 \\
Silhouette & Layer size & 0.1104 & 0.0470 \\
Silhouette & Number layers & 0.1596 & 0.0922 \\
Silhouette & Input order & 0.0387 & 0.0112 \\
Silhouette & Number labels & 0.3206 & 0.3158 \\
 \bottomrule
\end{tabular}
\caption{Reuters statistics of experiment groups.}
\label{tab:reuters-statistics}
\end{table}

\clearpage
\medskip

\bibliography{references}